\documentclass[pdflatex,sn-mathphys-num]{sn-jnl}

\usepackage{graphicx}%
\usepackage{float}%
\usepackage{multirow}%
\usepackage{amsmath,amssymb,amsfonts}%
\usepackage{amsthm}%
\usepackage{mathrsfs}%
\usepackage[title]{appendix}%
\usepackage{xcolor}%
\usepackage{textcomp}%
\usepackage{manyfoot}%
\usepackage{booktabs}%
\usepackage{algorithm}%
\usepackage{algorithmicx}%
\usepackage{algpseudocode}%
\usepackage{listings}%

\theoremstyle{thmstyleone}%
\theoremstyle{thmstyletwo}%

\theoremstyle{thmstylethree}%

\begin{document}

\title[Article Title]{Exploring AI in Fashion: A Review of Aesthetics, Personalization, Virtual Try-On, and Forecasting}


\author{\fnm{Laila} \sur{Khalid}}\email{laila@mail.ustc.edu.cn}

\author*{\fnm{Wei} \sur{Gong}}\email{weigong@ustc.edu.cn}

\affil{\orgdiv{Department of Computer Science}, \orgname{University of Science and Technology of China}, \orgaddress{\street{96, JinZhai Road}, \city{Hefei}, \postcode{230026}, \state{Anhui}, \country{China}}}

\abstract{Fashion-focused artificial intelligence has rapidly advanced in recent years, driven by deep learning and its deployment in recommender systems, detection, retrieval, and analytics. Yet several consumer-facing domains remain comparatively under-surveyed despite their practical impact. This work provides a comprehensive review of methods, datasets, and evaluation metrics across four such domains: aesthetics, personalization, virtual try-on, and forecasting. We synthesize technical approaches spanning representation learning, preference modeling, image transformation, and time-series analysis; relate them to downstream recommender systems and user experience; and highlight cross-domain dependencies (e.g., aesthetics-informed personalization, trend-informed recommendations). We also catalog commonly used datasets and metrics, including those from object detection and image segmentation pipelines, where relevant to try-on and visual understanding. Finally, we identify open challenges and promising directions for integrated AI-driven fashion systems.}




\maketitle

\section{Introduction}\label{sec1}

The fashion industry has undergone a dramatic transformation in recent years, with trends emerging and becoming obsolete within days. The integration of Internet shopping has brought unprecedented revenue growth to this sector, creating exciting opportunities for analyzing visual features in clothing. These visual features, encompassing color, texture, pattern, silhouette, and style elements, form the foundation of fashion design and directly influence individual consumer preferences and purchasing decisions. Today's artificial intelligence algorithms have evolved beyond simple product recommendations to genuinely evaluate the compatibility and aesthetics of fashion items~\cite{R61, R58}, effectively making AI systems primary curators of style rather than human experts alone.

Recent research has explored diverse approaches to fashion AI, spanning implicit user feedback systems~\cite{R07}, scenario-oriented recommendations~\cite{R129}, probabilistic multimedia ontologies~\cite{R132}, hidden Markov models for behavior characterization~\cite{R133}, and collaborative filtering with visual attention mechanisms~\cite{R134}. Other work has focused on curating compatible garments~\cite{han2017learning, hsiao2017learning, R38}, employing Generative Adversarial Networks for clothing synthesis~\cite{shih2018compatibility}, or recommending complete outfit combinations~\cite{R79}. Studies have also examined item interchangeability and compatibility confidence~\cite{R39, R40}, human perception of beauty and aesthetics~\cite{R51, R52, R53, R54}, and fashion assessment in social networks~\cite{SimoSerraCVPR2015}.

Despite these rapid advances, existing surveys have predominantly organized the field around technical tasks, such as detection, segmentation, retrieval, and synthesis, rather than around the functional objectives that these technologies serve. While technically oriented taxonomies are valuable for understanding algorithmic approaches, they obscure a more fundamental question: \textit{How do AI systems collectively support the end-to-end fashion consumption experience?} We argue that a more insightful organization emerges when we consider fashion AI through the lens of the consumer decision-making journey and the closed-loop system that supports it.

This survey introduces a novel organizing framework centered on four interconnected application domains: \textbf{aesthetics}, \textbf{personalization}, \textbf{virtual try-on}, and \textbf{fashion forecasting}. These domains were not selected arbitrarily but represent the essential stages of a complete fashion AI ecosystem that mirrors and supports consumer behavior:

\begin{enumerate}
    \item \textbf{Aesthetics} forms the foundational layer, addressing how AI systems learn and evaluate visual appeal, style coherence, and compatibility. Before any recommendation can be made, systems must understand what constitutes ``good'' fashion, the implicit rules governing color harmony, style consistency, and visual balance that make outfits appealing.
    
    \item \textbf{Personalization} builds upon aesthetic understanding by tailoring these principles to individual users. While aesthetics captures universal or culturally-shared notions of style, personalization recognizes that fashion is inherently subjective, what appeals to one consumer may not appeal to another. This stage translates general aesthetic knowledge into individual preference models.
    
    \item \textbf{Virtual try-on} serves as the validation and visualization layer, enabling consumers to assess personalized recommendations on their own bodies before purchase. This domain addresses the critical gap between digital browsing and physical reality, reducing uncertainty and enabling informed decisions.
    
    \item \textbf{Fashion forecasting} closes the loop by predicting future trends, informing what styles and items should enter the recommendation pipeline. Forecasting ensures that aesthetic models and personalization systems remain current, while also helping consumers consider the longevity and relevance of their choices.
\end{enumerate}

Together, these four domains form a \textbf{closed-loop system}: forecasting anticipates what will be desirable, aesthetics evaluates whether items meet quality and compatibility standards, personalization matches items to individual users, and virtual try-on validates choices, with consumer feedback and market outcomes cycling back to inform future forecasts. This interdependence means that advances in one domain directly enable or constrain progress in others, making an integrated understanding essential for researchers and practitioners alike.

This survey fills a gap in the literature not merely by covering under-explored topics, but by providing a principled framework that reveals how disparate research threads connect to form a functioning fashion AI system. Previous surveys, including the comprehensive work by Cheng et al.~\cite{cheng2021fashion}, have catalogued technical approaches within recommendation, analysis, detection, and synthesis. Our contribution differs in its organizing logic: rather than asking ``what techniques exist?'', we ask ``what functions must a complete fashion AI system perform, and how do they interrelate?'' This functional perspective offers researchers a roadmap for identifying bottlenecks, understanding dependencies, and prioritizing future work.

Our study compiles highly-referenced and influential papers published between 2013 and 2025, organized as follows: Section~\ref{sec2} reviews related surveys and positions our contribution. Section~\ref{sec3} covers aesthetics, the foundational understanding of visual appeal and style principles. Section~\ref{sec4} discusses personalization methods that tailor experiences to individual users. Section~\ref{sec5} presents virtual try-on technologies that simulate how clothing appears on users. Section~\ref{sec6} provides an overview of forecasting methods for predicting future trends. Section~\ref{sec7} outlines future directions, and Section~\ref{sec8} presents concluding remarks.

\section{Related Work and Comparative Analysis}\label{sec2}

In this section, we position our survey within the broader landscape of fashion AI research, comparing our approach with existing surveys and highlighting the distinctive nature of our selected domains.

\subsection{Previous Fashion AI Surveys}

\begin{table*}[h!]
\centering
\caption{Prior Relevant Survey Works In Fashion}\label{tab1}
\begin{tabular}{@{}p{4cm} p{3cm} p{6cm}@{}}
\hline
\textbf{Work} & \textbf{Focus} & \textbf{Key Highlights} \\
\hline
Cheng et al. \cite{cheng2021fashion} & Recommendation, Style Analysis, Synthesis, Detection & 200 pivotal works, offering a thorough overview of intelligent fashion breakthroughs spanning detection, analysis, synthesis, and recommendation. \\
\hline
Chen et al. \cite{chen2023survey} & Fashion analysis, Recommendation, Synthesis, Virtual Try-On & Explores cutting-edge techniques such as makeup transfer and virtual try-on, outlining future directions. \\
\hline
Islam et al. \cite{10.1145/3636552} & Image Retrieval & Heightened attention on Fashion Image Retrieval (FIR). Proposes future extensions to include diverse fashion items like watches, body ornaments, and accessories. \\
\hline
Ding et al. \cite{10.1145/3627100} & Recommendation, Personalized recommendation & Fashion recommendation in computational fashion research, addressing personalized product, mix-and-match, and outfit recommendations. \\

\hline
Giri et al. \cite{8763948} & Supply chain & AI methods, such as machine learning and image recognition, at various supply chain stages and business perspectives (B2B/B2C). \\
\hline
Chakraborty et al. \cite{informatics8030049} & Recommendation Systems & Examines existing studies, offering insights into methods, algorithmic models, and filtering techniques. \\
\hline
Jain et al. \cite{article223} & Fashion Trend Analysis & Exploration of novel concepts like fashion creativity and cultural inclusivity in the fashion domain. \\
\hline
Ghodhbani et al. \cite{article445} & Virtual Try-On & Review outlines challenges such as shape, pose, and texture, providing a comprehensive overview of research advancements. \\
\hline
Song et al. \cite{song2024image} & Virtual Try-On & First systematic image-based virtual try-on survey with unified evaluation, covering pipeline architectures, person representation, and clothing warping strategies. \\
\hline
\end{tabular}
\end{table*}

Fashion AI has garnered significant attention from researchers, resulting in several survey papers that examine specific technological aspects. Cheng et al. \cite{cheng2021fashion} provided a thorough overview of intelligent fashion breakthroughs spanning detection, analysis, synthesis, and recommendation, cataloging 200 pivotal works. While comprehensive, this survey did not examine the interconnections between these domains or their collective impact on the consumer experience. A more compact version of this can be seen in this study by Chen et al. \cite{chen2023survey}.

Image retrieval in fashion contexts was extensively covered by Islam et al. \cite{10.1145/3636552}, examining techniques for finding visually similar or complementary fashion items. This survey concentrated on Fashion Image Retrieval (FIR) techniques rather than the subjective aspects of fashion aesthetics or personalization.

Supply chain applications of AI were explored by Giri et al. \cite{8763948}, focusing on machine learning and image recognition at various supply chain stages from business-to-business and business-to-consumer perspectives. This work addressed industry operations rather than consumer-facing technologies.

\begin{figure*}[h!]
\centering
\includegraphics[width=\textwidth]{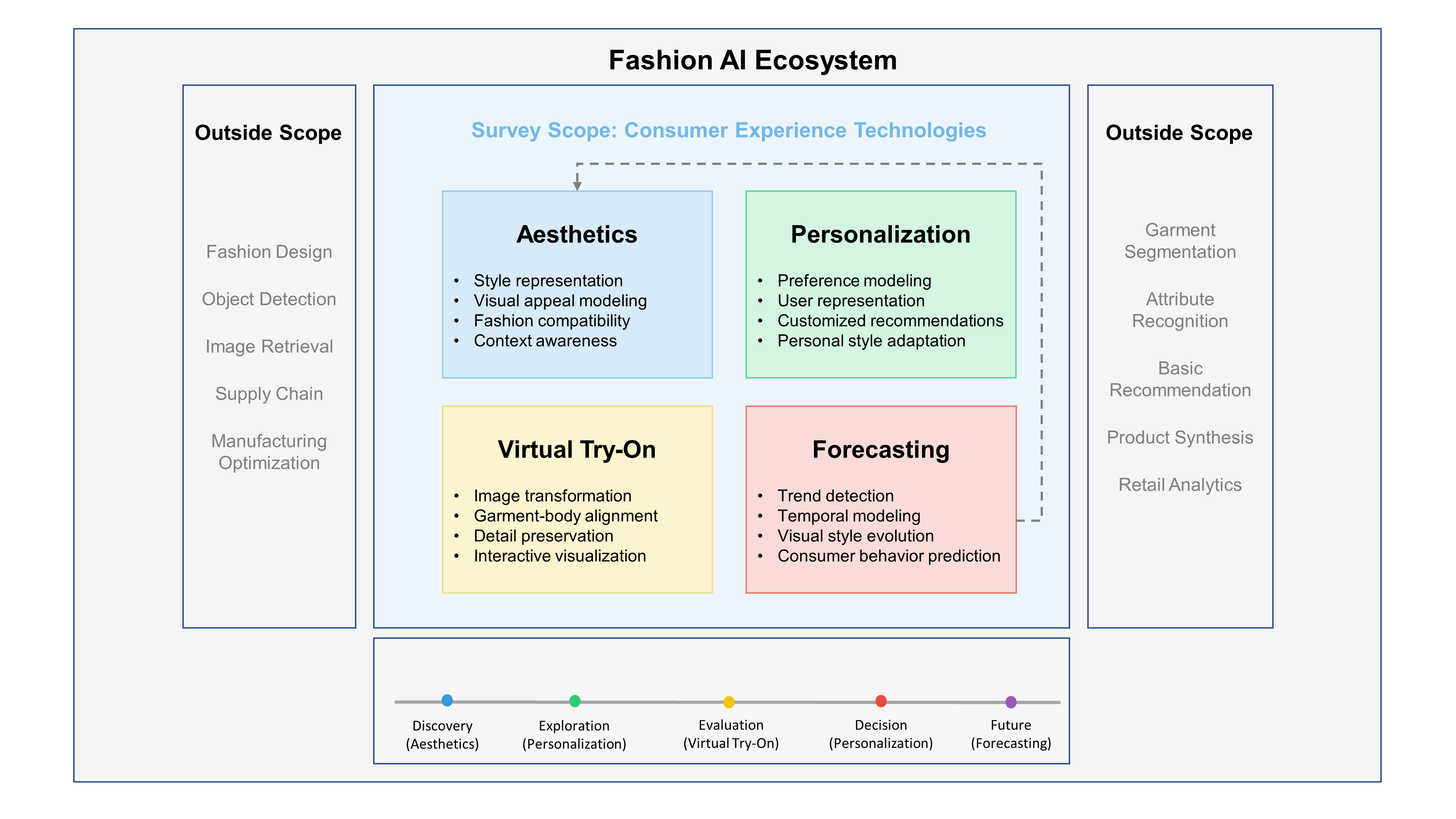}
\caption{Integrated Fashion AI Framework showing domain interconnections and information flow between aesthetics, personalization, virtual try-on, and forecasting systems.}
\label{fig:domain_flow}
\end{figure*}

Recommendation systems received dedicated attention in \cite{informatics8030049} and \cite{10.1145/3627100}, offering insights into methods, algorithmic models, and filtering techniques. These surveys primarily focused on the technical aspects of recommendation without broader integration with aesthetic principles or visualization technologies.

Fashion trend analysis was examined by Jain et al. \cite{article223}, exploring novel concepts like fashion creativity and cultural inclusivity. While related to our forecasting domain, this work did not connect trend analysis to other consumer-facing technologies.

Virtual try-on technologies were reviewed by Ghodhbani et al. \cite{article445}, outlining challenges such as shape, pose, and texture while providing an overview of research advancements. Similarly, Chen et al. \cite{chen2023survey} also explore cutting-edge techniques, including virtual try-on. Work by Song et al. \cite{song2024image} provides the first systematic image-based virtual try-on survey with unified evaluation. However, this work focuses specifically on the technical aspects of virtual try-on methods rather than examining their integration with other fashion AI domains or their role in consumer-facing applications.

In summary, while existing surveys have made valuable contributions by deeply examining specific technological aspects of fashion AI, they have typically treated these domains as separate technical challenges. Our work differs fundamentally by examining aesthetics, personalization, virtual try-on, and forecasting as an interconnected ecosystem that collectively shapes the consumer fashion experience. This approach enables us to identify cross-domain synergies and technical opportunities that would be missed in a more isolated analysis.

\subsection{Integrated Framework for Fashion AI Domains}

To address the limitations of existing surveys, we propose an integrated framework that connects the four core domains of fashion AI. This framework demonstrates how aesthetics, personalization, virtual try-on, and forecasting work together to create a comprehensive consumer experience.

As illustrated in Figure~\ref{fig:domain_flow}, our selected domains exhibit technical independence while maintaining application interdependence. Each domain employs distinct technical approaches:

\begin{itemize}
    \item \textbf{Aesthetics} primarily employs visual perception models and style representation learning
    
    \item \textbf{Personalization} utilizes preference learning and adaptive recommendation systems
    
    \item \textbf{Virtual Try-On} focuses on image transformation, garment physics, and synthesis techniques
    
    \item \textbf{Forecasting} leverages time series analysis and trend detection methodologies
\end{itemize}

\subsection{Cross-Domain Integration Analysis}

The four domains of fashion AI—aesthetics, personalization, virtual try-on, and forecasting—exhibit significant interdependencies that create both opportunities and challenges for integrated system development.

\begin{table*}[h!]
\centering
\caption{Cross-Domain Integration Analysis: How Each Domain Addresses Gaps in Others}\label{tab:cross-domain-integration}
\small
\begin{tabular}{@{}p{2.5cm} p{4cm} p{6cm}@{}}
\hline
\textbf{Domain} & \textbf{Key Challenges} & \textbf{How Other Domains Help} \\
\hline
Aesthetics & Subjective evaluation, context dependency & \textbf{Personalization:} User preference modeling \\
& & \textbf{Virtual Try-on:} Real-time visual feedback \\
& & \textbf{Forecasting:} Trend-aware aesthetics \\
\hline
Personalization & Cold-start problem, preference drift & \textbf{Aesthetics:} Universal beauty principles \\
& & \textbf{Virtual Try-on:} Interactive preference learning \\
& & \textbf{Forecasting:} Trend-based personalization \\
\hline
Virtual Try-On & Realism, pose diversity, garment fit & \textbf{Aesthetics:} Visual quality assessment \\
& & \textbf{Personalization:} User-specific adaptation \\
& & \textbf{Forecasting:} Trend-aware garment selection \\
\hline
Forecasting & Cold-start for new products, external signal integration & \textbf{Aesthetics:} Visual trend analysis \\
& & \textbf{Personalization:} User behavior patterns \\
& & \textbf{Virtual Try-on:} User interaction data \\
\hline
\end{tabular}
\normalsize
\end{table*}

\begin{figure*}[h!]
\centering
\includegraphics[width=\textwidth]{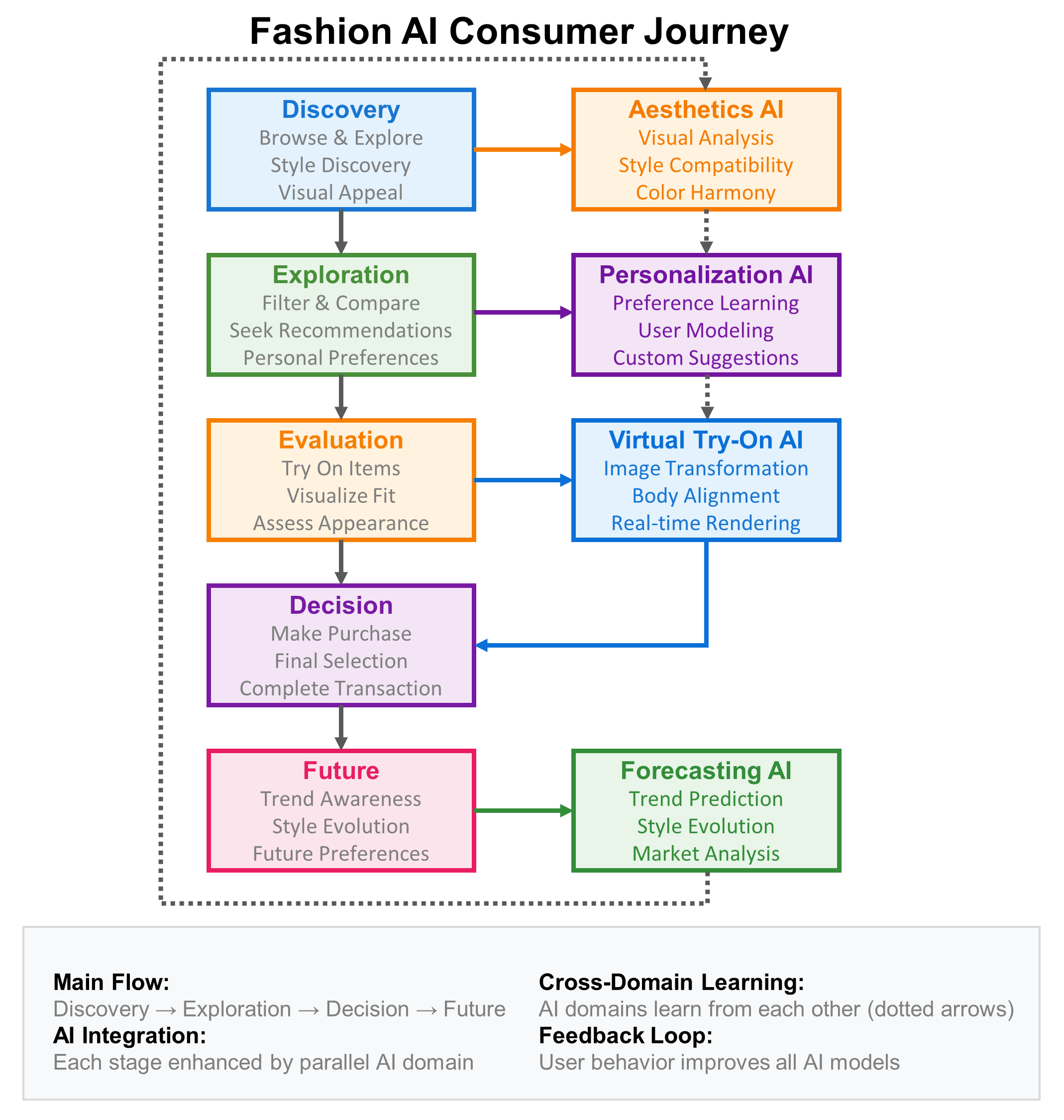}
\caption{Technical Architecture of Fashion AI Domains showing direct connections between aesthetics, personalization, virtual try-on, and forecasting systems with cross-domain learning pathways.}
\label{fig:domain_architecture}
\end{figure*}

This technical diversity means each domain can be (and often has been) studied in isolation. However, when deployed together, these domains create an integrated system with circular influence patterns where outputs from one domain become inputs for another (Figure~\ref{fig:domain_architecture}).

\section{Aesthetics In Fashion}\label{sec3}
\begin{figure*}[h!]
\centering
\includegraphics[width=\textwidth]{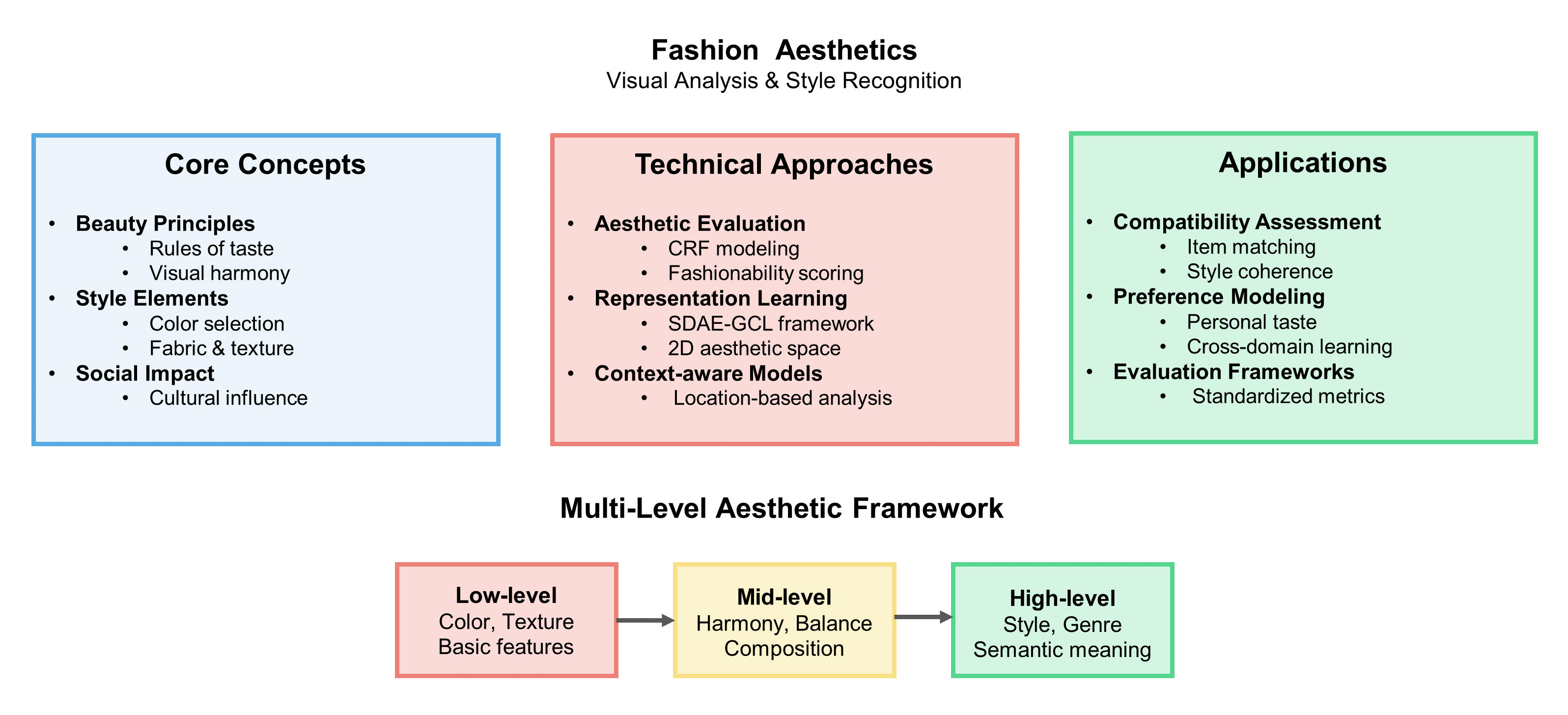}
\caption{Fashion Aesthetics framework for visual analysis and style recognition.}
\label{fig:aesthetics}
\end{figure*}

Aesthetics in fashion serves as the foundational layer that influences all other AI domains, establishing universal principles of visual appeal, style compatibility, and aesthetic judgment that inform personalization, virtual try-on, and forecasting systems. The aesthetic principles developed here create the visual language that enables effective personalization, realistic virtual try-on experiences, and accurate trend prediction. This section examines core technical approaches, including aesthetic evaluation methods, representation learning for aesthetics, and context-aware aesthetic modeling; explores learning paradigms from supervised methods to cross-domain knowledge transfer; and analyzes application domains covering compatibility assessment, consumer preference modeling, and aesthetic evaluation frameworks. The framework illustrated in Figure~\ref{fig:aesthetics} demonstrates how aesthetic principles are captured and applied across the fashion AI ecosystem.

\subsection{Core Technical Approaches}

Aesthetics in fashion represents a foundational element that shapes both design principles and consumer perception. According to The Stanford Encyclopedia of Philosophy \cite{sep-aesthetic-concept}, the term “aesthetic” originated in the 18th century, referring to objects, judgments, attitudes, experiences, or values related to beauty. The concept typically emerges from taste, with 18th-century theory emphasizing rationalist discourse on beauty and the influence of individualism on virtuosity. In the fashion domain, aesthetics provide essential building blocks—establishing rules of beauty, laws of taste, and artistic expression used by designers to create harmonious and appealing looks. They help define a designer's or brand's genre and flair, influencing color selection, fabric choices, and content direction, ultimately determining which audience a design appeals to.

Aesthetic evaluation approaches have advanced significantly with the evolution of computer vision technology. Neuroaesthetics \cite{SimoSerraCVPR2015} made a notable contribution by examining various fashionability factors, including outfit type and gender, garment type, user and setting characteristics, and fashionability scoring. Their Conditional Random Field (CRF) model and bag-of-words approach provided comprehensive feedback based on these combined factors. Their analysis of 144,169 user posts containing images, text, and metadata established valuable resources for subsequent research. The importance of aesthetic evaluation is underscored by Magic Closet \cite{liu2012hi}, who determined that individuals wearing well-coordinated, visually appealing attire were perceived as having higher status, demonstrating the social impact of fashion aesthetics.

Representation learning for aesthetics has seen innovative approaches that capture complex visual properties. VF-ISS-AWS (Visual Features-Image Scale Space-Aesthetic Weighted Space) \cite{jia2016learning} is a framework that positions aesthetics in a continuous image-scale space (2D space) as an intermediate level of representation. This structure facilitates higher-level perception of aesthetic effects. They employed a Stacked Denoising Autoencoder Guided by Correlative Labels (SDAE-GCL), effectively mapping visual features into this image scale space using both labeled and unlabeled datasets. The SDAE-GCL's denoising optimization mechanism removes random noise from initial inputs, improving accuracy and experimental reliability. Another approach, DCFA (Dynamic Collaborative Filtering model with Aesthetic Features) \cite{yu2018aesthetic}, recognized that different individuals express different aesthetic characteristics and recommended tensor factorization models for personalizing aesthetic attributes to ensure comprehensive aesthetic coverage.

Context-aware aesthetic modeling represents an important direction, acknowledging that fashion aesthetics are situational. The Trip Outfits Advisor \cite{zhang2017trip} pioneered an investigation into relationships between location attributes and apparel attributes, developing recommendations for location-oriented apparel in online travel photos. This work highlighted that wearing fashionable outfits in inappropriate contexts can undermine their aesthetic value. Similar contextual awareness is demonstrated in DeepStyle \cite{tautkute2019deepstyle}, which presented an innovative end-to-end method using a neural network architecture to model joint multi-modal space representation, where visual and linguistic signals are transformed together. This approach addresses the challenge of formulating natural queries for specific fashion searches across multiple modalities, such as “this dress but silk made." FHSI (Fuzzy Perceptual Color Model) \cite{shamoi2022color} examined the context dependency of color aesthetics, specifically focusing on color harmonies and color impressions in art and fashion domains. This work introduced a novel fuzzy perceptual color model that aligns with human perception and enables effective modeling of aesthetic judgments. Community-based aesthetic learning has emerged as a novel approach for large-scale aesthetic preference modeling. PMTM (Probabilistic Multi-Topic Model) \cite{Probabilistic} segments clients into aesthetic quality communities based on shared dress color preferences. The model employs geometry-based feature selection to capture aesthetic tendencies using visual, semantic, and textual features in a 457-dimensional representation. The system identifies tightly connected aesthetic communities through graph-based frameworks and dense graph mining techniques, achieving 85.7\% accuracy on a dataset of over one million clients across 18 aesthetic communities.

\subsection{Learning Paradigms}

Fashion aesthetic modeling has employed diverse learning approaches to capture the subjective and multifaceted nature of aesthetic judgment.

Supervised learning methods have been widely applied to aesthetic modeling tasks. Traditional approaches focused on training matching principles—for instance, pairing a red t-shirt with white jeans rather than green ones. More sophisticated supervised approaches like the one used by Neuroaesthetics \cite{SimoSerraCVPR2015} employed CRF to model fashionability factors. FARM (Fashion Aesthetic Recommendation Model) \cite{lin2019improving} is an improved framework incorporating supervised generation loss to better encode aesthetic information. This approach illustrates the challenge of balancing multiple objectives, as generation quality may divert attention from recommendation performance, creating a trade-off between visual aesthetic enhancement and effective recommendation.

Feature extraction and fusion techniques provide alternative approaches to aesthetic learning. Several studies have examined recommending fashion garments that complete incomplete outfits \cite{han2017learning, R38}. Earlier methods primarily focused on selecting garments from databases that coordinate well with query items \cite{he2016monomer, hsiao2017learning, R38}. The hybrid multilabel Convolutional Neural Network (CNN) and Support Vector Machine (SVM) (mCNN-SVM) approach in Trip Outfits Advisor \cite{zhang2017trip} demonstrates how multiple techniques can be combined to learn associations between clothes and contextual factors. These methods excel at capturing specific aspects of aesthetics but may struggle with integrating broader fashion principles.

Cross-domain knowledge transfer represents an advanced paradigm for aesthetic learning. ACDN (Aesthetic Cross-Domain Networks) \cite{liu2020exploiting} introduced a deep network that shares parameters characterizing personal aesthetic preferences and transfers knowledge between domains. This approach leverages an aesthetic network to extract features and integrates them into a cross-domain network for transferring domain-independent preferences. Network cross-connections enable dual knowledge transfer across domains. This paradigm recognizes that aesthetic preferences often transcend specific fashion categories and contexts. Taking standardization further, the A100 framework \cite{9879586} presents Aesthetic 100 as an industry standard with systematic principles and guidelines for performance testing at both characteristic and general levels.

FHSI \cite{shamoi2022color} employed a fuzzy-based learning approach to investigate the context dependency of aesthetic principles. They integrated fuzzy set theory with color space modeling to create the fuzzy perceptual color model, a novel color representation framework that accommodates the inherent imprecision and subjectivity of human color perception. Their experimental methodology compared aesthetic evaluations across different contexts (art and fashion) to determine which aesthetic principles remained consistent and which varied. Their findings revealed that color harmony—a fundamental aspect of aesthetic appeal—exhibited a high degree of universality. This suggests that despite contextual differences, certain fundamental aesthetic principles remain consistent across domains.

Weakly-supervised community learning represents an advanced paradigm for aesthetic modeling. PMTM \cite{Probabilistic} employs weak supervision to link client tags with aesthetic community memberships, using Jensen-Shannon divergence to measure similarity between aesthetic profiles. The approach constructs affinity graphs reflecting client relationships and applies specialized algorithms to identify tightly connected subgroups defined by shared dress color preferences. This method addresses the challenge of modeling subjective aesthetic preferences at scale while maintaining individual variation within broader aesthetic communities.

\subsection{Application Domains}

Aesthetic principles in fashion influence multiple application areas, each leveraging different aspects of visual appeal and harmony.

Compatibility assessment represents a fundamental application of fashion aesthetics. Previous research has predominantly addressed recognition issues, such as matching street scene items to in-store inventory \cite{R41, R42, R43, R44}. While this represents only a portion of the fashion recognition challenge, it directly benefits retailers. From a different perspective, recommendations can suggest interchangeable items rather than single garments that fit multiple outfits \cite{R39, R40}. These approaches indirectly incorporate aesthetic principles without explicitly addressing them. DeepStyle's \cite{tautkute2019deepstyle} Siamese network approach enables retrieval of items with similarity across multiple feature spaces from multimedia databases, supporting aesthetic-based matching across modalities.

Consumer preference modeling applies aesthetic principles to understand and predict individual choices. When appearance-oriented consumers purchase products, their decision criteria focus on whether items look good or satisfy aesthetic needs. ACDN \cite{liu2020exploiting} addressed this challenge by enhancing the modeling of individual inclinations from an aesthetic viewpoint through cross-domain recommendation methods. Their Feed-Forward Neural Network (FFNN) and Inception Local Global Network (ILGNet) approaches demonstrated improved performance in capturing personal aesthetic preferences. These applications recognize the subjective nature of aesthetic judgment while identifying patterns that can inform personalized recommendations.

Aesthetic evaluation frameworks provide standardized approaches to assessing fashion aesthetics. The A100 framework \cite{9879586} presents an industry standard (A100), which provides an elaborate characterization of AI models' aesthetic capabilities. This framework includes systematic aesthetic principles and guidelines for performance testing at both characteristic and general levels, establishing a definitive evaluation approach. The framework employs two multiple-choice examinations—LAT (Liberalism Aesthetic Test) and AAT (Academicism Aesthetic Test)—to address bottom-up and top-down fashion aesthetic criteria. Such frameworks help standardize aesthetic assessment in an inherently subjective domain.

FHSI \cite{shamoi2022color} expands application domains by providing tools for cross-context aesthetic evaluation. Their fuzzy color representation model enables automated extraction of color palettes that correspond to specific aesthetic impressions like “elegant," “formal," or “romantic." By examining the context-dependency of these aesthetic principles, their research provides valuable insights for applications spanning e-commerce, marketing, architectural and interior design, product design, fashion, web design, art, and merchandising. Their findings that color harmony is largely universal while impressions are partially context-dependent have significant implications for developing automated aesthetic assessment systems that can function effectively across different visual domains.

Aesthetic community identification represents a novel application domain for personalized recommendations. PMTM \cite{Probabilistic} demonstrated automatic identification of aesthetic quality communities through their probabilistic multi-topic model. The system segments clients into communities based on shared dress color preferences, enabling community-aware aesthetic assessment tools. This approach provides insights for e-commerce personalization and trend analysis, where understanding both individual and community-level aesthetic preferences is crucial for effective recommendation systems.

\begin{table*}
\centering
\caption{Taxonomy of Aesthetic Methods in Fashion: Technical Classification and Evaluation}\label{tab4}
\small
\begin{tabular}{@{}p{3cm} p{3cm} p{3cm} p{3cm}@{}}
\hline
\textbf{Work} & \textbf{Approach} & \textbf{Paradigm} & \textbf{Domain} \\
\hline
\multicolumn{4}{@{}l@{}}{\textit{Aesthetic Evaluation Approaches}} \\
\hline
Magic Closet \cite{liu2012hi} \newline(2012) & Status-based fashion evaluation & Supervised matching principles & Outfit recommendation with status impact \\
\hline
Neuroaesthetics \cite{SimoSerraCVPR2015} \newline(2015) & Multi-factor fashionability analysis & Conditional Random Field with bag-of-words & Comprehensive fashion evaluation \\
\hline
\multicolumn{4}{@{}l@{}}{\textit{Representation Learning for Aesthetics}} \\
\hline
VF-ISS-AWS \cite{jia2016learning} \newline(2016)& Continuous image-scale space mapping & Stacked Denoising Autoencoder (SDAE-GCL) & High-level aesthetic effect modeling \\
\hline
DCFA \cite{yu2018aesthetic} \newline(2018) & Brain-inspired aesthetic modeling & Deep structure with tensor factorization & Aesthetic-aware recommendation \\
\hline
FARM \cite{lin2019improving} \newline(2019)  & Visual-aesthetic information integration & Co-supervision learning with deep CNN & Fashion compatibility with aesthetic enhancement \\
\hline
\multicolumn{4}{@{}l@{}}{\textit{Context-Aware and Multi-Modal Aesthetics}} \\
\hline
Trip Outfit \cite{zhang2017trip}\newline (2017) & Location-apparel attribute association & Hybrid CNN-SVM with multi-label learning & Location-specific outfit recommendation \\
\hline
DeepStyle \cite{tautkute2019deepstyle} \newline(2019) & Multi-modal joint representation & Siamese network with feature fusion & Cross-modal fashion retrieval \\
\hline
ACDN \cite{liu2020exploiting} \newline(2020) & Aesthetic preference transfer & Cross-domain networks (ACDN) & Cross-domain recommendation \\
\hline
A100 \cite{9879586} \newline(2022)& Standardized aesthetic assessment & Systematic testing framework (A100) & Comprehensive aesthetic evaluation \\
\hline
FHSI \cite{shamoi2022color} \newline(2022) & Fuzzy perceptual color modeling & Fuzzy-based context comparison & Cross-domain aesthetic evaluation \\
\hline
PMTM \cite{Probabilistic} \newline(2025) & Community-based aesthetic learning with probabilistic multi-topic modeling & Weakly-supervised learning with graph-based clustering & Personalized aesthetic community identification \\
\hline
\end{tabular}
\normalsize
\end{table*}

\begin{table*}
\centering
\caption{Evaluation Methods and Datasets in Fashion Aesthetics Research}\label{tab4b}
\begin{tabular}{@{}p{3cm} p{2cm} p{3cm} p{4cm}@{}}
\hline
\textbf{Work} & \textbf{Metrics} & \textbf{Dataset} & \textbf{Key Contribution} \\
\hline
Magic Closet \cite{liu2012hi} & Status perception & Custom & Connecting visual outfit appeal to social status perception \\
\hline
Neuroaesthetics \cite{SimoSerraCVPR2015} & Accuracy, Precision, Recall, IOU, L1 & Fashion144k & Comprehensive fashionability factors analysis \\
\hline
VF-ISS-AWS \cite{jia2016learning} & MSE, MAE & Custom (Amazon, JD, Style, NWFW) & 2D continuous image-space for high-level aesthetics \\
\hline
Trip Outfit \cite{zhang2017trip} & Macro-F, mAP, AP, AUC & Fashionista, CCP, ColorfulFashion, Journey Outfit & Context-aware clothing recommendation for locations \\
\hline
DCFA \cite{yu2018aesthetic} & Recall, NDCG & Amazon Clothing & Neural network with tensor factorization for aesthetics \\
\hline
FARM \cite{lin2019improving} & AUC, MRR & FashionVC, ExpFashion & Layer-to-layer matching for aesthetic information \\
\hline
DeepStyle \cite{tautkute2019deepstyle} & Euclidean and L1, Cosine, Chi Squared & Polyvore & Joint representation of visual and linguistic signals \\
\hline
ACDN \cite{liu2020exploiting} & HR, NDCG, MRR & Amazon & Cross-domain aesthetic preference modeling \\
\hline
A100 \cite{9879586} & LAT and AAT tests & Maryland, FashionVC, Mytheresa, UTZappos50k & Standardized aesthetic capability testing framework \\
\hline
FHSI \cite{shamoi2022color} & Fuzzy similarity, Harmony score & Custom & Context-dependency analysis of color aesthetics \\
\hline
PMTM \cite{Probabilistic} & BER, F1, Precision, Recall & Custom (1M+ clients, 18 aesthetic communities) & Community-based aesthetic preference modeling with multi-modal features\\
\hline
\end{tabular}
\end{table*}

\subsection{Datasets and Evaluation}

The fashion aesthetics field lacks a single universally adopted dataset, with researchers typically using either hybrid datasets or custom collections. Many researchers create proprietary datasets by scraping websites like Amazon, JD, Style, and NWFW. The Fashionista Dataset, containing 158,235 images from chictopia.com obtained in 2011, represents a notable resource, though it lacks annotations and is stored in tab-delimited text files that must be combined to form a unified table. Other significant cross-domain datasets include Clothing Co-Parsing (CCP) and Colorful Fashion by Yang et al. \cite{yang2014clothing}, Journey Outfit Dataset by Trip Outfits Advisor \cite{zhang2017trip}, ExpFashion in NOR (Neural Outfit Recommendation) \cite{lin2019explainable}, Polyvore, released in Bidirectional Long Short-Term Memory (Bi-LSTM) \cite{han2017learning}, and Fashion144k, introduced in Neuroaesthetics \cite{SimoSerraCVPR2015}. FHSI \cite{shamoi2022color} employed custom datasets comprising 1,276 paintings by 10 artists from different movements and periods alongside 10,000 highly-liked fashion images for their investigation of color aesthetics across contexts.

Evaluation of aesthetic models primarily employs benchmark metrics including Mean Average Precision (mAP), average similarity score, Mean Reciprocal Rank (MRR), Area Under the Curve (AUC), Intersection over Union (IoU), Balanced Error Rate (BER), and L1-norm regularization. These metrics collectively provide comprehensive views of model precision and consistency in representing aesthetic qualities, demonstrating how effectively models capture and depict aesthetic preferences. Quantitative measures are sometimes combined with qualitative assessments from fashion experts or public feedback to form more holistic evaluations. This combination of quantitative and qualitative assessment creates a more complete picture of model quality and applicability.

\section{Personalization in Fashion}\label{sec4}
\begin{figure*}[h!]
\centering
\includegraphics[width=\textwidth]{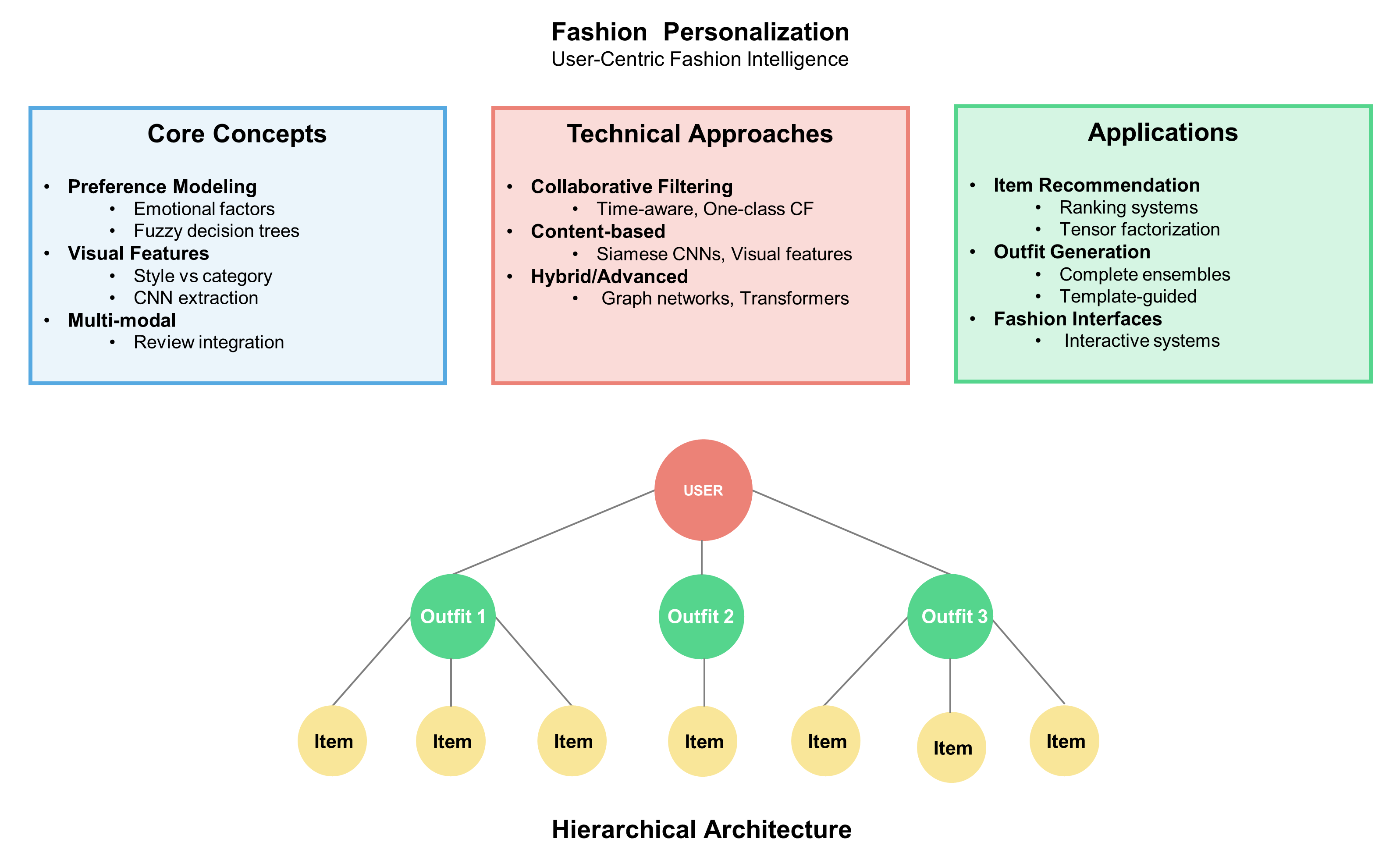}
\caption{Personalization framework for user-centric recommendations.}
\label{fig:personalization}
\end{figure*}

Personalization in fashion builds upon the aesthetic foundations established in the previous section to create tailored experiences that adapt to individual user preferences and behavioral patterns. This domain bridges universal aesthetic principles with individual taste, incorporating insights from both aesthetics and forecasting to deliver recommendations that are both visually appealing and personally relevant. This section explores core technical approaches, including preference modeling, user representation learning, and recommendation generation techniques; examines learning paradigms from collaborative filtering to hybrid methods; and analyzes application domains spanning individual item recommendation, outfit recommendation, and preference-aware interfaces. The framework presented in Figure~\ref{fig:personalization} demonstrates how these components work together to create adaptive, personalized fashion experiences.

\subsection{Core Technical Approaches}

Personalization has become a critical aspect of the fashion industry, enabling experiences tailored to individual preferences and needs. The global apparel market has shown remarkable growth, with projected revenue of 1.7 trillion dollars \cite{Gitnux_2023}, representing a significant year-on-year increase of 13.7 percent. This growth highlights the industry's resilience and potential for innovation, particularly in personalization technologies.

Preference modeling approaches form the foundation of fashion personalization systems. Wang et al. \cite{wang2014intelligent} demonstrated that emotional connections to fashion products can override rational considerations in purchase decisions. Their work utilized fuzzy decision trees to model these emotional factors alongside personalized body forms. More structured methods for capturing preferences include the functional tensor factorization approach FPITF (Functional Pairwise Interaction Tensor Factorization) \cite{hu2015collaborative}, which transforms multi-modal feature vectors of fashion items into a shared latent space where fashion-eye interactions are modeled. This method decomposes high-order feature vector interactions into pairwise combinations, enabling a more effective fashion item ranking. FashionDPO \cite{FashionDPO}, a direct preference optimization approach, fine-tunes diffusion models for fashion image generation by learning from human preference data. Their method addresses the challenge of aligning AI-generated fashion content with human aesthetic preferences through direct optimization of preference signals, enabling more personalized and aesthetically pleasing fashion image generation.

User representation learning has advanced significantly with deep neural networks showing promising results for image content representation, as demonstrated in seminal works by \cite{R25}, \cite{R26}, \cite{R27}, and \cite{R28}. A notable development in this area is DeepStyle \cite{ds2}, which conceptualizes fashion items as having two components: style and category. Their approach leverages CNN extracted visual information, removing category information to isolate style cues, thus creating more nuanced user preference profiles. VECF (Visually Explainable Collaborative Filtering) \cite{chen2019personalized} further enhanced user representation by developing an attention model that focuses on important regions in fashion images while incorporating user review information. More recently, MG-PFCM (Metapath-Guided Personalized Fashion Compatibility Modeling) \cite{guan2022personalized}, a multi-modal content-oriented user embedding module, was proposed that derives user embeddings based on the multi-modal contents of their interacted items, addressing the challenge of representing users with limited content information.

Recommendation generation techniques have evolved into sophisticated systems combining multiple factors. Many methods approach this challenge through either pairwise compatibility metrics, as seen in works by \cite{R39}, \cite{R57}, and \cite{R58}, or by modeling outfits as sets or ordered sequences. The latter approach is exemplified by Li et al. \cite{RO6}, who classified outfits as popular or unpopular, and the Bi-LSTM \cite{han2017learning} model for sequential outfit generation. Advanced recommendation approaches include the visually-aware model DVBPR (Deep Visually-aware Bayesian Personalized Ranking) \cite{kang2017visually}, which uses image content and Generative Adversarial Network (GAN) technology to generate personalized recommendations that match user preferences. TOG (Template-guided Outfit Generation) \cite{ding2023personalized} introduced the concept of “templates" (category combinations) to capture users' coordination preferences in fashion outfits, enabling more refined modeling of user preferences at both template and item levels. MG-PFCM \cite{guan2022personalized} is a metapath-guided approach that incorporates item attributes as an important entity type, addressing a gap in previous personalized fashion compatibility modeling methods. In a significant advancement, DiFashion \cite{xu2024diffusion}, a generative outfit recommender model, leverages diffusion models to actually generate new fashion items tailored to users' preferences, rather than merely recommending existing products. StyleMe \cite{wu2023styleme}, an AI-aided fashion design system, enables personalized clothing sketches and style transfer. The system employs a GAN framework with two components: a sketch generation model with adaptive channel feature normalization and channel attention modules to learn designer-specific styles, and a style transfer model that decouples content features from sketches and style features from reference images. PFNet \cite{sun2024pfnet} uniquely integrates attribute-aware personalized fashion editing with explainable compatibility analysis. It employs an unsupervised garment attribute decoupling network that uses hierarchical style control and mutual information minimization to independently encode attributes without requiring labeled data, addressing a significant limitation of previous approaches that relied on supervised information.

\begin{table*}
\centering
\caption{Taxonomy of Personalization Methods in Fashion: Technical Classification and Evaluation}\label{tab3}
\begin{tabular}{@{}p{2.5cm} p{3.5cm} p{3cm} p{4cm}@{}}
\hline
\textbf{Work} & \textbf{Approach} & \textbf{Paradigm} & \textbf{Domain} \\
\hline
\multicolumn{4}{@{}l@{}}{\textit{Preference Modeling Approaches}} \\
\hline
Wang et al. \cite{wang2014intelligent} \newline (2014) & Fuzzy decision trees modeling emotional factors & Knowledge-driven with design expertise & Preference-aware interface with body form consideration \\
\hline
 FPITF \cite{hu2015collaborative} \newline (2015) & Tensor factorization of multi-modal features & Collaborative filtering with latent space mapping & Individual item recommendation with ranking \\
\hline
 He et al. \cite{he2016ups} \newline (2016)& Time-dependent user preference modeling & One-class collaborative filtering & Individual item recommendation with trend analysis \\
\hline
 FashionDPO \cite{FashionDPO}\newline (2025) & Multi-expert feedback with DPO & Direct preference optimization with low rank adaptation & Personalized outfit generation with enhanced diversity \\
\hline
\multicolumn{4}{@{}l@{}}{\textit{Visual and Multi-modal Representation}} \\
\hline
DVBPR \cite{kang2017visually} \newline (2017)& Visually-aware preference modeling & Siamese CNNs with Bayesian ranking & Individual item recommendation with visual similarity \\
\hline
DeepStyle \cite{ds2} \newline (2017)& Style-category decomposition & CNN-based with feature disentanglement & Style representation with category awareness \\
\hline
VECF \cite{chen2019personalized} \newline (2019)& Region-aware attention modeling & Multimodal attention network with LSTM & Individual item recommendation with visual explanations \\
\hline
\multicolumn{4}{@{}l@{}}{\textit{Advanced Outfit and Ensemble Recommendation}} \\
\hline
 Bi-LSTM \cite{han2017learning} \newline (2017)& Sequential outfit modeling & Bidirectional LSTM & Complete outfit generation \\
\hline
 POG \cite{chen2019pog} \newline (2019)& Joint user-outfit preference integration & Transformer architecture & Outfit recommendation with user preference balancing \\
\hline
 FHN \cite{lu2019learning} \newline (2019)& Binary code learning for fast retrieval & Hashing network with set composition & Outfit recommendation with efficient matching \\
\hline
 HFGN \cite{li2020hierarchical} \newline (2020)& Hierarchical relationship modeling & Graph neural network & Outfit recommendation with user-item-outfit connections \\
\hline
 LPAE \cite{lu2021personalized} \newline (2021)& Anchored preference representation & Stacked self-attention with latent vectors & Outfit recommendation for cold-start users \\
\hline
MG-PFCM \cite{guan2022personalized} \newline (2022)& Metapath-guided heterogeneous graph & Graph attention with metapath-guided learning & Personalized compatibility modeling with attribute awareness \\
\hline
TOG \cite{ding2023personalized} \newline (2023)& Template-guided preference modeling & Joint learning with user-template-item interactions & Personalized outfit generation with compatibility balancing \\
\hline
StyleMe \cite{wu2023styleme} \newline (2023)& GAN-based dual-model framework with feature normalization & Hybrid generative modeling with style-content decoupling & Designer-specific sketch generation and style transfer \\
\hline
PFNet \cite{sun2024pfnet} \newline (2024)& Unsupervised attribute decoupling with compatibility-aware attention & Multi-task learning with hierarchical style control and global perception & Attribute-aware fashion editing with explainable compatibility analysis \\
\hline
DiFashion \cite{xu2024diffusion} \newline (2024)& Conditional diffusion model & Parallel generation with multi-condition guidance & Generative outfit recommendation with personalized image synthesis \\
\hline
\end{tabular}
\end{table*}

\subsection{Learning Paradigms}

The evolution of personalization systems in fashion has involved diverse learning approaches, each addressing different aspects of the personalization challenge.

Collaborative filtering methods have been widely adopted, with He et al. \cite{he2016ups} developing a theme-based, scalable model that incorporates both visual data (product photos) and textual data (user comments). Their one-class collaborative filtering approach accounts for time-dependent dynamics, providing insights into aesthetic characteristics and fashion trends. These methods excel at capturing patterns across user behaviors but often struggle with the “cold start" problem for new users or items.

Content-based approaches overcome some limitations of collaborative filtering by focusing on item attributes. DVBPR \cite{kang2017visually} utilized Siamese CNNs with Bayesian Personalized Ranking to build upon prior visually-aware recommendation systems. Their end-to-end learning strategy enables direct modeling of visual preference patterns. These approaches are particularly valuable in fashion, where visual attributes are paramount to user preferences.

Hybrid and advanced methods represent the current state-of-the-art in personalization learning. POG (Personalized Outfit Generation) \cite{chen2019pog} pioneered an industrial-scale approach that integrates user preferences for both individual items and complete outfits through a Transformer architecture. Taking a different approach, HFGN (Hierarchical Fashion Graph Network) \cite{li2020hierarchical} identified limitations in conventional compatibility matching approaches that fail to model the relationships among users, outfits, and items. They address this by constructing a hierarchical structure linking user-outfit interactions and outfit-item mappings. For users with limited data, the LPAE (Learnable Personalized Anchor Embedding) \cite{lu2021personalized}  framework uses anchored latent vectors (anchors) to represent user preferences in the outfit space, with item interactions captured through stacked self-attention. Another work, TOG \cite{ding2023personalized} jointly learns user-template interaction, user-item interaction, and outfit compatibility. MG-PFCM \cite{guan2022personalized} presents a metapath-guided heterogeneous graph learning approach to better capture the high-order relations among various entities (users, items, and attributes). Most recently, DiFashion \cite{xu2024diffusion}, a novel generative paradigm for outfit recommendation, was proposed, which utilizes diffusion models. This work features three specialized conditions (category prompt, mutual condition, and history condition) to guide the parallel generation of multiple fashion images that exhibit high fidelity, compatibility, and personalization, representing a significant shift from traditional retrieval-based methods. StyleMe \cite{wu2023styleme} is a dual-model GAN framework that addresses both personalization of sketch generation and style transfer for fashion design. This approach is notable for its data efficiency, requiring relatively few designer-specific sketches (around 100 per designer) while maintaining style consistency, making it practical for real-world fashion design applications where extensive designer-specific data may be limited. PFNet \cite{sun2024pfnet} introduces a multi-task fashion learning approach that combines unsupervised attribute disentanglement with explainable compatibility modeling. The method overcomes the limitation of requiring supervised attribute labels through a novel hierarchical style control mechanism with regularization constraints to reduce attribute correlation. FashionDPO \cite{FashionDPO} represents a significant advancement in preference-based learning paradigms by introducing direct preference optimization for fashion diffusion models. This approach learns directly from human preference comparisons rather than relying on traditional supervised learning signals, enabling the model to capture better aesthetic preferences and generate fashion content that better aligns with human taste through direct optimization of preference rankings.

\subsection{Application Domains}

Fashion personalization technologies have been applied across multiple domains, each addressing different aspects of the user experience.

Individual item recommendation represents the most common application domain. FPITF \cite{hu2015collaborative} proposes a system that ranks fashion items based on computed favorability scores derived from millions of pairwise feature vector combinations. This approach enables highly personalized single-item recommendations tailored to individual user preferences. The methods typically focus on finding items that match a user's established style preferences while potentially introducing novel elements that align with those preferences.

Outfit recommendation extends personalization beyond single items to complete ensembles. POG \cite{chen2019pog} made significant advances in this domain; the model simultaneously considers outfit coordination and individual user preferences. Similarly, FHN (Fashion Hashing Network) \cite{lu2019learning} collected user-labeled information and outfit data from fashion-focused social media to generate personalized outfit recommendations. These systems must balance aesthetic coherence with personalized style preferences—a more complex challenge than single-item recommendation. Extending personalization to outfit generation, TOG \cite{ding2023personalized} addressed the challenge of both personalizing and ensuring compatibility in newly generated outfits with their multi-step generation framework that first predicts templates based on user preferences. MG-PFCM \cite{guan2022personalized} focused on personalized fashion compatibility modeling, addressing whether a bottom (top) matches a given top (bottom) for a specific user. This approach uniquely incorporated attribute entities, which contain rich semantics that previous methods had overlooked in the compatibility estimation process. Taking outfit recommendation to a new level, DiFashion \cite{xu2024diffusion} introduced Generative Outfit Recommendation (GOR), which transcends the limitations of existing fashion products by generating entirely new fashion items that can be composed into visually compatible outfits tailored to users' unique fashion tastes.

Preference-aware fashion interfaces represent emerging applications that adapt the entire fashion shopping experience. While not explicitly detailed in the current literature, this domain encompasses systems that modify search results, browsing experiences, and visual presentations based on user preferences. These applications leverage the preference modeling and recommendation techniques to create holistic personalized experiences that extend beyond simple product recommendations. StyleMe \cite{wu2023styleme} extended preference-aware interfaces to the design process itself with an interactive environment that enables rapid generation of personalized sketches and exploration of diverse style options through reference-guided style transfer. Meanwhile, PFNet \cite{sun2024pfnet} is a novel application domain that bridges personalized fashion editing with compatibility analysis, enabling users to interactively edit clothing attributes while receiving real-time feedback on the compatibility of their designs with reference garments. This approach represents a significant advancement in preference-aware fashion interfaces by providing attribute-level compatibility explanations, helping users understand which specific aspects (color, pattern, style, shape) of their edited designs contribute to or detract from overall compatibility with other items. FashionDPO \cite{FashionDPO} extends preference-aware interfaces to generative fashion design through its direct preference optimization framework. The system enables users to generate personalized fashion images that align with their aesthetic preferences by learning directly from human preference data, creating a new paradigm for AI-assisted fashion design that prioritizes human taste and aesthetic judgment in the generation process.

\begin{table*}
\centering
\caption{Evaluation Methods and Datasets in Fashion Personalization Research}\label{tab3b}
\begin{tabular}{@{}p{2.5cm} p{3cm} p{3cm} p{4cm}@{}}
\hline
\textbf{Work} & \textbf{Metrics} & \textbf{Dataset} & \textbf{Key Contribution} \\
\hline
Wang et al. \cite{wang2014intelligent} & Human evaluation & CAD generated & Emotion-driven personalization with body form awareness \\
\hline
FPITF \cite{hu2015collaborative} & NDCG & Polyvore & Functional tensor factorization for user-item interaction modeling \\
\hline
He et al. \cite{he2016ups} & AUC & Amazon.com & Time-dependent dynamics for improving recommendation quality \\
\hline
DVBPR \cite{kang2017visually} & AUC & Amazon.com, Tradesy.com & End-to-end learning with Siamese-CNN architecture \\
\hline
Bi-LSTM \cite{han2017learning} & Compatibility prediction & Polyvore & Bidirectional LSTM for fashion compatibility learning \\
\hline
POG \cite{chen2019pog} & FITB, Compatibility Prediction (CP) & Taobao's iFashion Dataset  & Transformer-based encoder-decoder for personalized outfit generation \\
\hline
FHN \cite{lu2019learning} & AUC, NDCG, FITB & Polyvore Dataset & Hashing method for efficient personalized set recommendation \\
\hline
VECF \cite{chen2019personalized} & F1, HR, NDCG & Amazon.com & Attention modeling for important visual regions with user reviews \\
\hline
HFGN \cite{li2020hierarchical} & HR, NDCG, Recall, Precision & Taobao's iFashion Dataset & Graph-based modeling of user-outfit-item relationships \\
\hline
LPAE \cite{lu2021personalized} & AUC, NDCG & Polyvore & Anchor-based representation for new users with limited data \\
\hline
MG-PFCM \cite{guan2022personalized} & AUC, MRR & IQON3000 & Metapath-guided graph learning for attribute-aware compatibility modeling \\
\hline
TOG \cite{ding2023personalized} & Precision, Recall, NDCG, CEE & iFashion, Polyvore-U & Template-guided preference learning for personalized outfit generation \\
\hline
StyleMe \cite{wu2023styleme} & FID, LPIPS, Human evaluation & Custom fashion dataset & Dual GAN framework for personalized sketch generation and style transfer \\
\hline
PFNet \cite{sun2024pfnet} & AUC, HR@K, User study & FashionVC, Polyvore-Maryland & Unsupervised attribute decoupling and explainable compatibility modeling with global perception \\
\hline
DiFashion \cite{xu2024diffusion} & FID, IS, Cosine Similarity, CIS, LPIPS, Human evaluation & iFashion, Polyvore-U & Diffusion-based generative outfit recommendation with multi-conditional guidance \\
\hline
FashionDPO \cite{FashionDPO} & IS, IS Accuracy, Compatibility, Personalization & iFashion, Polyvore-U & Multi-expert feedback with DPO framework \\
\hline
\end{tabular}
\end{table*}

\subsection{Datasets and Evaluation}

Several key datasets have emerged for evaluating personalized fashion recommendation systems. The Taobao iFashion Dataset, introduced in POG \cite{chen2019pog}, stands as the largest publicly available resource, containing 1.01 million outfits and 583K individual clothing items. The Polyvore dataset, released in Bi-LSTM \cite{han2017learning}, and the Amazon.com dataset also serve as important benchmarks in the field. The IQON3000 dataset, used by MG-PFCM \cite{guan2022personalized}, provides valuable evaluation resources with rich attribute information for each fashion item. DiFashion \cite{xu2024diffusion} utilized modified versions of the iFashion and Polyvore-U datasets specifically adapted for generative outfit recommendation tasks. These diverse datasets enable comprehensive evaluation across different recommendation scenarios and user populations.

The effectiveness of personalization models is assessed through various established metrics. Overall performance is commonly measured using AUC and ROC (Receiver Operating Characteristic) curve, which evaluate the models' prediction and discrimination abilities for individual users. More specific aspects of recommendation quality are evaluated using Normalized Discounted Cumulative Gain (NDCG), which measures ranking quality; Fill-in-the-Blank (FITB), which tests the ability to complete outfits; F1 score, which balances precision and recall; and Hit Ratio (HR), which measures the frequency of relevant recommendations. TOG \cite{ding2023personalized} introduced Compatibility Evaluation Experts (CEE) to quantitatively evaluate the compatibility of generated outfits, moving beyond the limitations of human evaluation or single-perspective metrics. MG-PFCM \cite{guan2022personalized} employed MRR to assess the effectiveness of complementary item retrieval in personalized fashion compatibility modeling. DiFashion \cite{xu2024diffusion} introduced a comprehensive evaluation framework for generative models, incorporating metrics for image fidelity (Fréchet Inception Distance (FID), Inception Score (IS), CLIP Score), similarity (Learned Perceptual Image Patch Similarity (LPIPS), CLIP Image Score (CIS)), compatibility (compatibility evaluator), and personalization.

\section{Virtual Try-On}\label{sec5}
\begin{figure*}[h!]
\centering
\includegraphics[width=\textwidth]{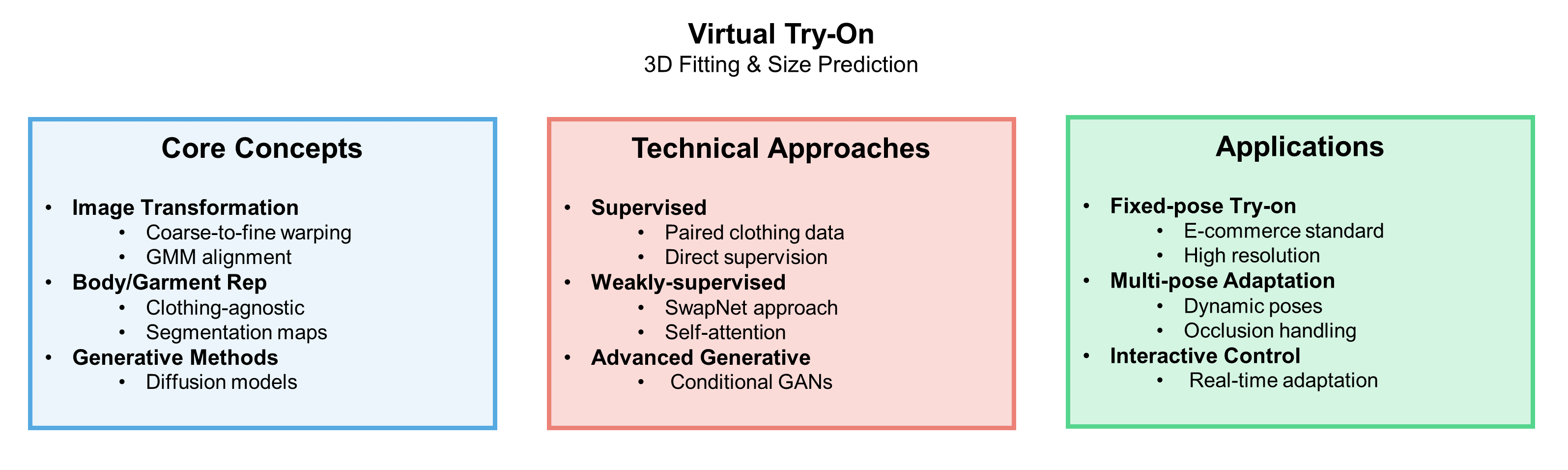}
\caption{Virtual Try-On framework for 3D fitting and size prediction.}
\label{fig:virtual_tryon}
\end{figure*}

Virtual try-on technology represents a critical bridge between aesthetic assessment and user decision-making in the fashion consumer journey, leveraging both the aesthetic principles and personalization insights developed in previous sections. As users progress from discovery and exploration to evaluation, virtual try-on systems provide visual feedback necessary for informed purchasing decisions while incorporating personalized preferences. This section examines core technical approaches spanning image transformation, body representation, and generative enhancement methods; explores learning paradigms from supervised to advanced generative approaches; and analyzes application domains including fixed-pose, multi-pose, and interactive try-on systems. The framework illustrated in Figure~\ref{fig:virtual_tryon} demonstrates how these technical components integrate to create seamless user experiences.

\subsection{Core Technical Approaches}

Virtual try-on technology has evolved through several distinct technical approaches, each addressing specific challenges in realistic garment rendering and transformation.

Image transformation techniques form the foundation of many virtual try-on systems. The pioneering work VITON (Virtual Try-On Network) \cite{han2018viton} introduced a coarse-to-fine warping framework that seamlessly transfers clothing items onto person images. This approach uses a multi-task encoder-decoder network to create initial visualizations, followed by refinement stages. Building on this foundation, CP-VTON (Characteristic-Preserving Virtual Try-On Network) \cite{wang2018toward}  introduced the Geometric Matching Module (GMM) to better align garments with body shapes while preserving clothing characteristics. More recent advancements include the SD-VITON (Sequential Deformation VITON) \cite{shim2024towards} framework, which addresses texture squeezing artifacts through disentangled appearance flow prediction, and the Progressive Inference Paradigm in PG-VTON (Progressive Generation VITON) \cite{fang2024pg}, which separates the try-on process into distinct specialized modules for improved performance.

Body and garment representation approaches have significantly improved the realism of virtual try-on systems. Initially, representations were simplistic, but VITON-HD (VITON High Definition) \cite{choi2021viton} introduced clothing-agnostic person representation, allowing the model to reduce dependency on the original clothing worn by the person. VTNFP (Virtual Try-on Network with Feature Preservation) \cite{yu2019vtnfp} employs a three-stage design that generates detailed body segmentation maps to delineate both body parts and clothing regions. Addressing specific anatomical challenges, CP-VTON+ (Characteristic-Preserving VTON Plus) \cite{minar2020cp} features enhanced skin labeling to better handle neck and chest areas previously misidentified as background. The semantic understanding of clothing-body relationships was further advanced by ACGPN (Adaptive Content Generating and Preserving Network) \cite{yang2020towards}, which explicitly models the spatial layout to preserve occlusion relationships naturally.

Generative enhancement methods represent the cutting edge of virtual try-on technology. Recent works have incorporated sophisticated generative techniques to improve visual quality. C-VTON (Context-driven VITON) \cite{fele2022c} employs conditional normalization layers, leveraging contextual information to guide the generation process. StableVITON \cite{kim2024stableviton} adapts pre-trained diffusion models for try-on tasks, using zero cross-attention blocks to condition intermediate features. Skin texture realism has been enhanced through specialized approaches like StyleGAN2-based inpainting in PG-VTON \cite{fang2024pg}, allowing for more natural transitions between garments and body parts. Recent advances in diffusion-based virtual try-on include OOTDiffusion (Outfitting Fusion based Diffusion) \cite{ootdiffusion}, which introduced an innovative outfitting fusion architecture that eliminates redundant warping processes through self-attention layers, providing precise alignment between garment features and target human bodies while incorporating outfitting dropout for enhanced controllability. FITMI \cite{fitmi} integrates virtual try-on with recommendation systems, providing personalized try-on experiences through Latent Diffusion Model (LDM)-based approaches with text inversion networks and automatic dataset selection capabilities. MV-VTON (Multi-View Virtual Try-On Network) \cite{MV-VTON} further advanced this paradigm by introducing multi-view virtual try-on capabilities through view-adaptive selection mechanisms and joint attention blocks for pose-aware clothing feature extraction, enabling comprehensive try-on experiences across different viewing angles.

\begin{table*}
\centering
\caption{Taxonomy of Virtual Try-On Methods: Technical Classification and Evaluation}\label{tab2}
\begin{tabular}{@{}p{3cm} p{3cm} p{3cm} p{4cm}@{}}
\hline
\textbf{Work} & \textbf{Approach} & \textbf{Paradigm} & \textbf{Domain} \\
\hline
\multicolumn{4}{@{}l@{}}{\textit{Image Transformation Techniques}} \\
\hline
VITON \cite{han2018viton} \newline (2018)& Coarse-to-fine warping with multi-task encoder-decoder & Supervised with refinement network & Fixed-pose with frontal-view try-on \\
\hline
CP-VTON \cite{wang2018toward} \newline (2018)& Geometric Matching Module (GMM) with composition mask & Supervised with cloth characteristic preservation & Fixed-pose with improved detail retention \\
\hline
 LA-VITON \cite{lee2019viton} \newline (2019)& Enhanced GMM with grid interval consistency loss & Supervised with occlusion handling & Fixed-pose with improved alignment \\
\hline
SD-VITON \cite{shim2024towards} \newline (2024)& Sequential Deformation framework with disentangled flow prediction & Adversarial with specialized TV loss & High-resolution with anti-squeezing mechanism \\
\hline
\multicolumn{4}{@{}l@{}}{\textit{Body and Garment Representation}} \\
\hline
VTNFP \cite{yu2019vtnfp} \newline (2019)& Three-stage design with body segmentation mapping & Self-attention enhanced with conditional GAN & Fixed-pose with improved body part preservation \\
\hline
CP-VTON+ \cite{minar2020cp} \newline (2020)& Enhanced GMM with explicit skin labeling & Supervised with improved segmentation & Fixed-pose with better neck/chest handling \\
\hline
ACGPN \cite{yang2020towards} \newline (2020)& Semantic layout preservation with adaptive generation & Second-order constraint on Thin-Plate Spline & Fixed-pose with occlusion relationship modeling \\
\hline
VITON-HD \cite{choi2021viton} \newline (2021)& Clothing-agnostic person representation & Misalignment-aware normalization & High-resolution synthesis (1024×768) \\
\hline
\multicolumn{4}{@{}l@{}}{\textit{Multi-Pose and Advanced Generation}} \\
\hline
MG-VTON \cite{dong2019towards} \newline (2019)& Multi-stage pose-guided appearance transformation & Conditional parsing with Warp GAN & Multi-pose adaptation with large variations \\
\hline
C-VTON \cite{fele2022c} \newline (2022)& Context-driven image generation & Conditional normalization layers & Context-aware try-on with environmental cues \\
\hline
PG-VTON \cite{fang2024pg} \newline (2024)& Progressive inference paradigm with three specialized modules & StyleGAN2-based skin inpainting & Fixed-pose with advanced skin preservation \\
\hline
Wear-Any-Way \cite{chen2024wear} \newline (2024)& Dual U-Net with sparse correspondence alignment & Interactive point-based control mechanism & User-controllable try-on with style manipulation \\
\hline
StableVITON \cite{kim2024stableviton} \newline (2024)& Zero cross-attention blocks for feature conditioning & Diffusion model adaptation & Fixed-pose with fine detail preservation \\
\hline
OOTDiffusion \cite{ootdiffusion} \newline (2025)& Outfitting fusion with self-attention alignment & Latent diffusion with outfitting dropout & Fixed-pose with enhanced controllability \\
\hline
FITMI \cite{fitmi} \newline (2025)& LDM with text inversion and mask-aware skip connections & Latent diffusion with recommendation integration & Commercial application with user experience focus \\
\hline
MV-VTON \cite{MV-VTON} \newline (2025)& View-adaptive selection with joint attention & Diffusion-based multi-view generation & Multi-view try-on with pose adaptation \\
\hline
\end{tabular}
\end{table*}

\subsection{Learning Paradigms}

The evolution of virtual try-on systems has been characterized by increasingly sophisticated learning approaches that balance supervision requirements with generation quality.

Supervised approaches dominated early virtual try-on systems, relying on paired data of clothing items and person images. VITON \cite{han2018viton} employs a multi-task encoder-decoder network trained with direct supervision to generate initial try-on results. CP-VTON \cite{wang2018toward} introduced composition mask learning to better control which parts of the aligned clothing should be preserved in the final synthesized image. LA-VITON (Looking-Attractive VITON) \cite{lee2019viton} enhanced geometric matching through grid interval consistency loss, demonstrating how specialized loss functions can improve specific aspects of the try-on process. These supervised methods establish strong baselines but often require carefully prepared training data.

Weakly-supervised methods emerged to address limitations in training data availability and quality. SwapNet \cite{raj2018swapnet} pioneered this direction, transferring garment information across images with arbitrary clothing, body poses, and shapes without requiring ideal paired data. VTNFP \cite{yu2019vtnfp} incorporates self-attention mechanisms to make correlation matching more robust, enabling the model to better understand relationships between different image regions despite imperfect supervision. These approaches demonstrate greater flexibility in real-world scenarios where perfectly paired data may be unavailable.

Advanced generative methods represent the current state-of-the-art in learning approaches. While conditional GANs have been widely used for image synthesis in virtual try-on, recent work has introduced more sophisticated techniques. StableVITON \cite{kim2024stableviton} uses attention total variation loss to produce sharper attention maps for more precise preservation of clothing details. SD-VITON \cite{shim2024towards} developed a disentangled appearance flow prediction that separates the optimization objectives for different aspects of the transformation process. Wear-Any-Way \cite{chen2024wear} introduced sparse correspondence alignment that enables precise user control over garment appearance. OOTDiffusion \cite{ootdiffusion} further advanced this paradigm with outfitting fusion that combines latent diffusion with self-attention mechanisms for improved controllability and alignment precision, while FITMI \cite{fitmi} introduced commercial-focused learning paradigms that integrate latent diffusion models with recommendation systems for personalized user experiences, and MV-VTON \cite{MV-VTON} developed multi-view learning paradigms that adapt clothing features based on person pose through view-adaptive selection mechanisms with hard-selection and soft-selection processes for global and local feature extraction.

\subsection{Application Domains}

Virtual try-on technology has expanded across several application domains, each addressing different user needs and technical challenges.

Fixed-pose try-on represents the foundational application, where clothing is transferred onto person images in standard poses—typically frontal views. Early VITON approaches focused on this scenario due to its relative simplicity. Significant progress has been made in high-resolution synthesis, with VITON-HD \cite{choi2021viton} achieving impressive results at 1024×768 resolution. Within this domain, attention has increasingly focused on detail preservation techniques for accurately rendering logos, patterns, and textures that are critical for realistic visualization. While constrained in pose variation, fixed-pose try-on systems offer practical value for e-commerce applications where standardized product presentation is common.

Multi-pose adaptation extends virtual try-on capabilities to handle diverse body positions and viewing angles. MG-VTON \cite{dong2019towards} addresses this challenge by disentangling the warping of clothes' appearance from pose manipulation. This approach handles large pose variations by processing these aspects in multiple distinct stages. Multi-pose systems must address complex challenges, including self-occlusions and significant garment deformations that occur when the body is in non-standard positions. These systems provide more versatile try-on experiences that better reflect real-world usage scenarios where clothing appearance changes substantially with body movement.

Interactive try-on systems represent the frontier of virtual try-on applications, offering unprecedented user control. Wear-Any-Way \cite{chen2024wear} revolutionized this domain with precise manipulation of how garments are worn through a point-based control system. Their dual U-Net architecture with sparse correspondence alignment allows users to interactively adjust features like sleeve rolling, coat opening, and tucking styles. This transition from passive visualization to interactive manipulation marks a significant advancement in user experience, moving virtual try-on systems closer to the flexibility of physical fitting rooms. Commercial virtual try-on systems represent a significant application domain, exemplified by FITMI \cite{fitmi}, which integrates latent diffusion models with recommendation systems, providing automatic dataset selection and customized preprocessing pipelines with advanced pose processing techniques and recommendation system integration, demonstrating the practical application of virtual try-on technology in commercial settings. MV-VTON \cite{MV-VTON} addresses multi-view applications in online shopping, enabling customers to see try-on results from multiple angles through their system with view-adaptive selection mechanisms that handle complex pose scenarios. OOTDiffusion \cite{ootdiffusion} contributes to e-commerce applications through its outfitting fusion approach, providing enhanced controllability for fashion retail and online shopping platforms with personalized garment recommendation systems.

\begin{table*}
\centering
\caption{Evaluation Methods and Datasets in Virtual Try-On Research}\label{tab2b}
\begin{tabular}{@{}p{3cm} p{3cm} p{3cm} p{3.5cm}@{}}
\hline
\textbf{Work} & \textbf{Metrics} & \textbf{Dataset} & \textbf{Key Contribution} \\
\hline
VITON \cite{han2018viton} & IS & VITON & First image-based virtual try-on framework \\

\hline
LA-VITON \cite{lee2019viton} & IS, SSIM & VITON & Improved geometric matching accuracy \\
\hline
VTNFP \cite{yu2019vtnfp} & A/B test & VITON & Three-stage design for accurate detail capture \\
\hline
MG-VTON \cite{dong2019towards} & A/B test, SSIM, IS & MPV, DeepFashion & Multi-pose handling with self-occlusion resolution \\
\hline
CP-VTON+ \cite{minar2020cp} & IoU, SSIM, LPIPS & VITON & Neck and chest area preservation \\
\hline
ACGPN \cite{yang2020towards} & SSIM, IS & VITON, CPVITON & Adaptive content preservation based on pose \\
\hline
VITON-HD \cite{choi2021viton} & FID, LPIPS & VITON-HD & First high-resolution (1024×768) approach \\
\hline
C-VTON \cite{fele2022c} & SSIM, FID & VITON & Context-guided synthesis for improved realism \\
\hline
SD-VITON \cite{shim2024towards} & FID, KID, LPIPS, SSIM & VITON-HD & Resolution of texture squeezing artifacts \\
\hline
PG-VTON \cite{fang2024pg} & FID, IS, hyperIQA, SSIM & VITON & Top-down inference pipeline for robust try-on \\
\hline
Wear-Any-Way \cite{chen2024wear} & FID, KID, SSIM, LPIPS & Custom, VITON-HD & First manipulable try-on with interactive control \\
\hline
StableVITON \cite{kim2024stableviton} & FID, KID, SSIM, LPIPS & VITON-HD, DressCode & End-to-end try-on with pre-trained diffusion models \\
\hline
OOTDiffusion \cite{ootdiffusion} & FID, KID, SSIM, LPIPS & VITON-HD, DressCode & Outfitting fusion with self-attention alignment \\
\hline
FITMI \cite{fitmi} & FID, IS, User experience metrics & DressCode, VITON-HD & Commercial virtual try-on with recommendation system \\
\hline
MV-VTON \cite{MV-VTON} & SSIM, LPIPS, FID, KID & MVG, VITON-HD, DressCode & Multi-view virtual try-on with pose adaptation \\
\hline
\end{tabular}
\end{table*}

\subsection{Datasets and Evaluation}

There are multiple datasets on the horizon now, but most of the works leverage the VITON dataset \cite{han2018viton}. It comprises 16,253 frontal-view women and top clothing image pairs. For high-resolution usage, VITON-HD \cite{choi2021viton} is available with an image size of 1024×768 comprising 13,679 frontal-view women and top clothing image pairs. Another noticeably famous dataset is the MPV dataset, introduced in MG-VTON \cite{dong2019towards}, which contains 35,687 person images and 13,524 clothes images. Each person's image in MPV has different poses. The image is in the resolution of 256 × 192. DeepFashion \cite{liuLQWTcvpr16DeepFashion} is also a famous one that is frequently utilized. It contains over 800,000 diverse fashion images ranging from shop images to unconstrained consumer photos, constituting the largest visual fashion analysis database. Each image in this dataset is labeled with 50 categories, 1,000 descriptive attributes, a bounding box, and clothing landmarks. Others include DressCode \cite{morelli2022dress}, Digital Wardrobe \cite{bhatnagar2019mgn}, TailorNet Dataset \cite{patel20tailornet}, CLOTH3D \cite{bertiche2020cloth3d}, 3DPeople \cite{pumarola20193dpeople}, THUman Dataset \cite{tao2021function4d}, FashionOn \cite{hsieh2019fashionon} and VVT dataset \cite{dong2019fw}.

The evaluation metrics used to measure the performance of existing fashion try-on model modules include IS, which estimates realism and diversity of generated images. Other commonly used measures include the Structural Similarity Index (SSIM), VGG-based Perceptual Distance (PD), Perceptual Distance to Target Clothing (PD(TC)), Total Variation (TV), pairwise A/B testing, Hyper Image Quality Assessment (hyperIQA), Kernel Inception Distance (KID), and the L1 loss function. In recent works, we see LPIPS used in the paired setting, and FID adopted in the unpaired setting.

\section{Fashion Forecasting}\label{sec6}
\begin{figure*}[h!]
\centering
\includegraphics[width=\textwidth]{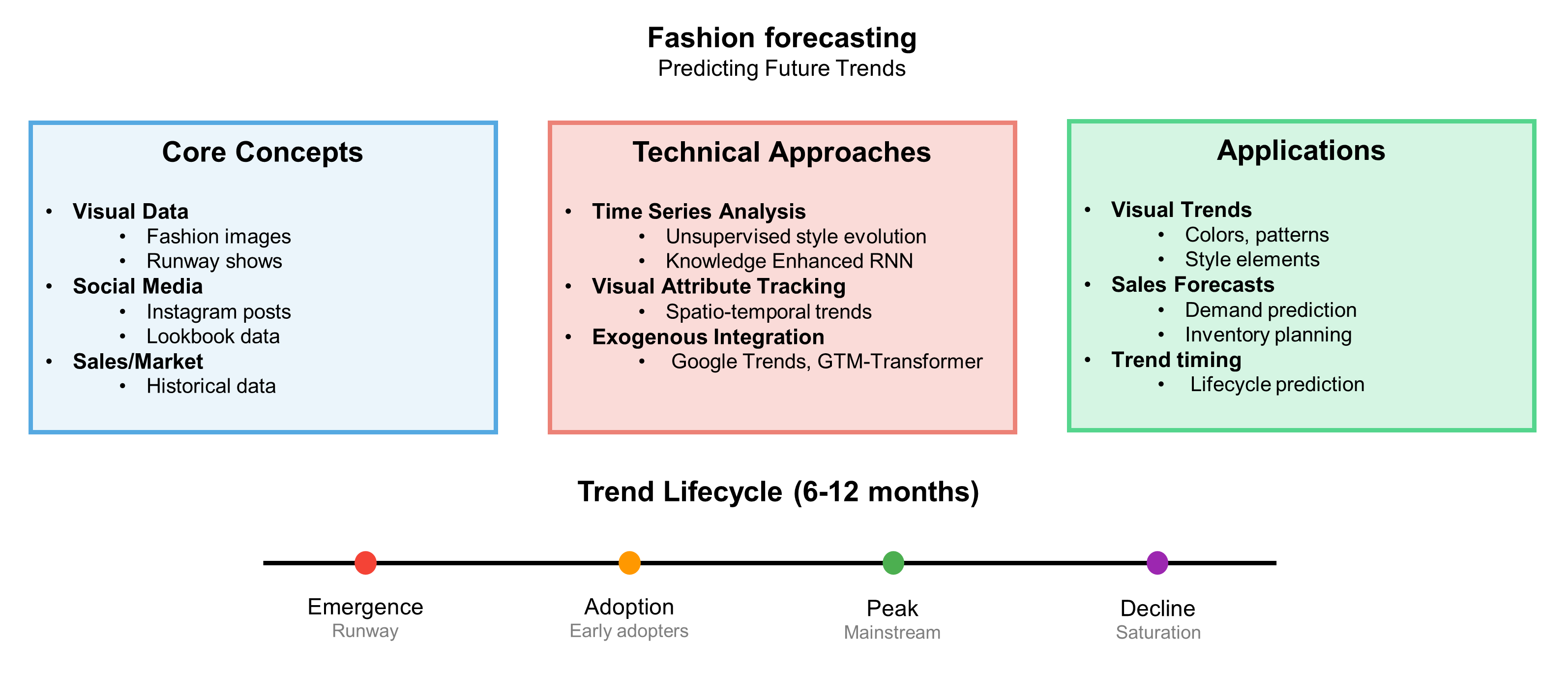}
\caption{Forecasting framework for trend prediction and market analysis.}
\label{fig:forecasting}
\end{figure*}

Fashion forecasting represents the forward-looking dimension of fashion AI, synthesizing insights from aesthetics, personalization, and virtual try-on to predict future trends and market dynamics. This domain completes the fashion AI ecosystem by providing temporal context that informs all other domains, enabling trend-aware personalization, future-relevant virtual try-on options, and aesthetic evolution prediction. This section explores core technical approaches, including time series analysis, visual attribute analysis, and sales and consumer behavior modeling; examines learning paradigms from unsupervised methods to transformer-based approaches; and analyzes application domains covering visual trend prediction, commercial performance forecasting, and social media trend analysis. The framework presented in Figure~\ref{fig:forecasting} demonstrates how these components integrate to create predictive capabilities that enhance the entire fashion AI system.

\subsection{Core Technical Approaches}

Fashion forecasting involves predicting emerging trends and styles that will dominate the industry in the near future using AI technologies. These approaches analyze extensive data on current fashion trends, consumer behavior, social media activity, and economic indicators using sophisticated algorithms and ML models.

Time series analysis forms a foundational approach in fashion forecasting, capturing the temporal evolution of styles and trends. Fashion Forward \cite{R89} pioneered an unsupervised approach to predicting future style popularity from fashion images. Their model demonstrates capabilities for imagining new style combinations that will gain popularity, identifying style dynamics (distinguishing between trendy and classic elements), and identifying key visual attributes that will dominate future fashion landscapes. Building on this foundation, KERN (Knowledge Enhanced Recurrent Network) \cite{ma2020knowledge} was developed as a tool for modeling time series data using a deep Recurrent Neural Network (RNN) architecture. This approach integrates various internal and external knowledge from the fashion domain alongside technical capabilities, creating a comprehensive system that accounts for the industry's unique challenges, including the rapid pace of change and multiple influential factors.

Visual attribute analysis represents another critical technical approach in fashion forecasting. The GeoStyle framework \cite{mall2019geostyle} automatically identifies clothing and style attributes in images and analyzes their spatial and temporal trends. Their framework demonstrated more accurate long-term trend forecasts for numerous fashion attributes and styles compared to prior approaches. This technique involves extracting visual features from fashion images and tracking their evolution across time and geographic regions. Similarly, Lo et al. \cite{lo2019dressing} developed methods relying on social media data of individual outfit looks, predicting popularity with high accuracy by analyzing visual elements.

Sales and consumer behavior modeling approaches complement visual analysis by incorporating market data. Loureiro et al. \cite{loureiro2018exploring} developed a deep learning approach to predict fashion industry sales and forecast future performance of new products. This method acknowledges that, despite fashion's short product lifecycle, inventory and purchase strategies can be informed by historical data from retailers' databases. These approaches focus less on visual elements and more on market performance patterns, connecting trend analysis to business outcomes. More recently, Giri et al. \cite{giri2022deep} proposed an intelligent forecasting system that combines image feature attributes extracted using deep learning with sales data to predict future demand for new fashion products. Their model uses machine learning clustering to identify product groups based on sales patterns and image similarity, then employs neural networks to classify new items and predict their sales profiles by finding the best match in established clusters. Sousa et al. \cite{censored} developed a two-stage approach to forecast demand for new products using censored data, employing Random Forest, Deep Neural Networks, and Support Vector Regression algorithms to predict demand for new products, with an additional model that weighted demands from historically similar products.

Exogenous knowledge integration has emerged as a promising approach in fashion forecasting, as demonstrated by GTM-Transformer (Google Trends Multimodal Transformer) \cite{skenderi2024well} , which introduced a non-autoregressive transformer model that systematically probes Google Trends time series data and combines it with multimodal product information to forecast sales of new fashion items. This model encodes Google Trends data representing product attribute popularity, then uses cross-attention to understand which portions of these exogenous signals are most relevant to the sales forecast. This approach effectively addresses the cold-start problem of forecasting for brand-new products without sales history, showing that the most useful exogenous information comes from 7-10 months prior to the product's planned release, typically the end of the previous year's corresponding fashion season.

\begin{table*}
\centering
\caption{Taxonomy of Fashion Forecasting Methods: Technical Classification and Evaluation}\label{tab5}
\begin{tabular}{@{}p{3cm} p{3cm} p{3cm} p{3cm}@{}}
\hline
\textbf{Work} & \textbf{Approach} & \textbf{Paradigm} & \textbf{Domain} \\
\hline
\multicolumn{4}{@{}l@{}}{\textit{Time Series Analysis Approaches}} \\
\hline
Fashion Forward \cite{R89} (2017) & Visual style evolution modeling & Unsupervised fashion image analysis & Visual trend prediction \\
\hline
KERN \cite{ma2020knowledge} \newline (2020)& Knowledge enhanced recurrent network & Deep RNNs with domain knowledge integration & Fine-grained fashion element trends \\
\hline
Neo-Fashion \cite{zhao2021neo} \newline (2021)& Periodic mode drift prediction & Walk model analysis with big data & Macroscopic fashion season forecasting \\
\hline
\multicolumn{4}{@{}l@{}}{\textit{Visual Attribute Analysis}} \\
\hline
Lo et al. \cite{lo2019dressing} \newline (2019)& Social media outfit analysis & Paper Doll Parsing with LSTM & Outfit popularity prediction \\
\hline
GeoStyle \cite{mall2019geostyle} \newline (2019)& Spatio-temporal attribute tracking & Parametric model with Gaussian mixture & Long-term style attribute forecasting \\
\hline
Chakraborty et al. \cite{chakraborty2020predicting} (2020)& Fashion week image analysis & Logistic regression & Runway-to-retail trend prediction \\
\hline
\multicolumn{4}{@{}l@{}}{\textit{Sales and Consumer Behavior Modeling}} \\
\hline
Loureiro et al. \cite{loureiro2018exploring} \newline (2018)& Historical sales pattern analysis & Deep learning with multiple metrics & Commercial performance forecasting \\
\hline
Giri et al. \cite{giri2022deep} \newline (2022)& Image features with sales data integration & Hybrid: unsupervised clustering with neural network classification & New product demand forecasting \\
\hline
GTM-Transformer \cite{skenderi2024well} \newline (2024)& Exogenous knowledge integration with multimodal data & Non-autoregressive transformer & First-order inventory optimization \\
\hline
Sousa et al. \cite{censored} \newline (2024)& Censored demand modeling with Expectation-Maximization (EM) & Two-stage methodology: EM for demand estimation + ML for forecasting & New product demand forecasting with substitute analysis \\
\hline
\end{tabular}
\end{table*}

\subsection{Learning Paradigms}

Fashion forecasting employs diverse learning methodologies to capture the complex dynamics of trend evolution and consumer preferences.

Unsupervised learning approaches offer advantages in discovering patterns without pre-defined categories. Fashion Forward \cite{R89} demonstrated the effectiveness of unsupervised methods in forecasting style popularity from fashion images, allowing the model to identify emergent patterns rather than being constrained by predetermined trend categories. This paradigm is particularly valuable in fashion, where novel styles may not fit existing classifications. Unsupervised approaches can uncover unexpected connections and emergent trends that supervised methods might miss.

Supervised prediction models provide alternative approaches when labeled data is available. Chakraborty et al. \cite{chakraborty2020predicting} proposed an approach using logistic regression to predict patterns and outfits based on Instagram posts from New York Fashion Week. Similarly, Lo et al. \cite{lo2019dressing} utilized Paper Doll Parsing models, InceptionV3, and LSTM architectures in a supervised manner to predict outfit popularity. These approaches leverage labeled datasets to train models that can predict specific aspects of future fashion trends based on historical patterns. Giri et al. \cite{giri2022deep} demonstrated the effectiveness of combining supervised classification with unsupervised clustering to categorize fashion items based on their visual features and sales profiles, using neural networks to achieve 72.4\% classification accuracy in predicting which sales pattern cluster a new item would follow. Sousa et al. \cite{censored} employed a two-stage supervised learning approach where the first stage transforms censored sales data into true demand estimates, while the second stage applies supervised learning algorithms (Random Forest, Deep Neural Networks, Support Vector Regression) to predict demand for new products.

Knowledge-enhanced methods integrate domain expertise with data-driven approaches. KERN \cite{ma2020knowledge} represents a sophisticated example, combining deep recurrent neural networks with fashion domain knowledge. This integration enhances the model's ability to account for industry-specific factors that purely data-driven approaches might miss. Similarly, Neo-Fashion \cite{zhao2021neo} presented a periodic mode drift prediction scheme using walk model analysis for information-driven fashion trend forecasting, incorporating domain understanding of fashion seasons. These approaches recognize that effective fashion forecasting requires both technical capabilities and industry knowledge.

Transformer-based methodologies have recently been applied to fashion forecasting, with GTM-Transformer \cite{skenderi2024well} demonstrating its effectiveness through a non-autoregressive transformer architecture. Unlike autoregressive approaches that can suffer from error compounding, their model generates the entire sales forecast at once by learning effective representations of both the product and exogenous data sources. The authors show that this non-autoregressive design provides a 4.4\% improvement in forecasting accuracy compared to equivalent autoregressive implementations, highlighting the advantages of transformers' parallel processing capabilities for fashion forecasting tasks.

\subsection{Application Domains}

Fashion forecasting technologies address various industry needs, each with specific focus and impact.

Visual trend prediction focuses on anticipating the evolution of style elements. According to theories of fashion innovation, new ideas typically originate in the designer's arena before undergoing trickle-down, trickle-across, and trickle-up processes. These ideas maintain relevance for at least one selling season, with successful trends lasting longer. Seasonal trends are heavily influenced by catwalk appearances, where they emerge as major items, colors, shapes, or styling approaches before declining over the subsequent 6-12 months as consumers adopt newer trends. The Neo-Fashion system \cite{zhao2021neo} addressed the complex problem of mode drift prediction at the macroscopic level for fashion seasons, moving beyond single-element prediction to consider colors, fabrics, styles, and intricate details collectively.

Commercial performance forecasting translates trend predictions into business implications. Loureiro et al.'s \cite{loureiro2018exploring} deep learning approach specifically targets sales prediction, helping businesses anticipate market performance. This application domain connects aesthetic and style trends to their commercial impact, supporting inventory planning, pricing strategies, and production decisions. Such forecasting helps businesses balance creative innovation with market viability, ensuring that products not only follow emerging trends but also meet consumer demand. Giri et al. \cite{giri2022deep} advanced this domain by developing a model capable of predicting weekly sales for new fashion items without historical data, addressing a critical industry challenge of inventory management for newly launched products. Building on this foundation, GTM-Transformer \cite{skenderi2024well} introduced a technique that directly addresses the “first-order problem" in fashion retail - determining optimal initial inventory quantities for new products before their market launch. Their approach demonstrated a 17.5\% reduction in monetary discrepancy compared to traditional ordering policies, potentially saving millions in inventory costs when scaled across a fashion retailer's product catalog, highlighting the substantial business impact of advanced forecasting technologies. Sousa et al. \cite{censored} focused specifically on forecasting demand for new products in fashion retailing, utilizing a comprehensive dataset from a European fashion retailer comprising 63.6 million records across 684 products from 2015-2016 bag collections.

Social media trend analysis leverages digital platforms' growing influence on fashion dissemination. Lo et al.'s \cite{lo2019dressing} study capitalizes on social media data to predict outfit popularity, recognizing these platforms as excellent vehicles for showcasing and circulating fashion trends to mass audiences. Similarly, Chakraborty et al.'s \cite{chakraborty2020predicting} approach analyzes Instagram posts from fashion events to predict future patterns. This application acknowledges that fashion runway shows provide inspiration for high-street and fast-fashion retailers, with social media accelerating trend dissemination and adoption.

\subsection{Datasets and Evaluation}

Fashion prediction relies heavily on diverse datasets, emphasizing variety due to the strong connection between prediction accuracy and consumer purchasing behavior. DeepFashion \cite{liuLQWTcvpr16DeepFashion} stands as one of the most popular datasets, offering comprehensive coverage with rich consumer data and versatile annotations for clothes detection, attribute prediction, fashion landmark estimation, and few-shot learning tasks. Social media-inspired datasets include Lookbook \cite{yoo2016pixelleveldomaintransfer}, Insta NYFW-19 dataset \cite{chakraborty2020predicting}, Neo-Fashion Dataset \cite{zhao2021neo}, and the StreetStyle dataset \cite{matzen2017streetstyle} from Instagram, and the Flickr 100M dataset \cite{thomee2016yfcc100m}. These resources capture the interplay between social media platforms and fashion runways. KERN \cite{ma2020knowledge} utilizes the Fashion Institute of Technology (FIT) dataset, which combines characteristics of institutional fashion knowledge with web-based data, merging the advantages of both knowledge and data-driven forecasting approaches. Giri et al. \cite{giri2022deep} utilized proprietary retail data from a European fashion retailer, comprising sales information and product images for 290 fashion items over a two-year period (2015-2016), demonstrating the value of industry collaborations for obtaining real-world commercial data.

Addressing the critical need for public datasets in new product sales forecasting, GTM-Transformer \cite{skenderi2024well} introduced VISUELLE, a comprehensive multimodal dataset containing 5,577 real fashion products from an Italian fast-fashion company. Each product in VISUELLE is associated with an image, textual metadata (category, color, fabric), 12 weeks of sales data, and three related Google Trends time series describing the popularity of product attributes. This enables benchmarking for new-product forecasting under realistic cold-start conditions.

Evaluation metrics for fashion forecasting focus primarily on prediction accuracy across various dimensions. Common metrics include Root Mean Square Error (RMSE), Mean Absolute Percentage Error (MAPE), Mean Absolute Error (MAE), and Mean Squared Error (MSE). For performance measurement, MAE and MAPE compare predictions against ground truth values to evaluate accuracy, while mAP assesses ranking performance. KL divergence provides insight into feature deviation from baselines, measuring how much each feature independently differs from expected patterns. MSE often serves as an objective function during training, while Cosine Similarity Error (CSE) indicates how closely predictions match actual ground truth patterns, reflecting the system's ability to capture trends correctly. For classification tasks within forecasting systems, metrics such as Classification Accuracy (CA), F1 score, precision, recall, and AUC of ROC curves help evaluate how well models categorize items into appropriate trend clusters or sales pattern groups. Recent advancements in evaluation methodology introduced in GTM-Transformer \cite{skenderi2024well} include the Weighted Absolute Percentage Error (WAPE), which expresses forecasting accuracy as a ratio of the sum of absolute errors to the sum of actual values, and the tracking signal (TS) measure, which quantifies forecasting bias by detecting systematic over- or underestimation. Additionally, the Edit Distance with Real Penalty (ERP) metric evaluates a model's ability to capture the actual dynamics of sales curves beyond simple point-wise accuracy. These expanded evaluation approaches provide a more comprehensive assessment of forecasting models, addressing both prediction accuracy and practical business utility.

\begin{table*}
\centering
\caption{Evaluation Methods and Datasets in Fashion Forecasting Research}\label{tab5b}
\begin{tabular}{@{}p{4cm} p{2cm} p{2cm} p{4cm}@{}}
\hline
\textbf{Work} & \textbf{Metrics} & \textbf{Dataset} & \textbf{Key Contribution} \\
\hline
Fashion Forward \cite{R89} & MAE, MAPE & DeepFashion & First unsupervised approach for visual style popularity prediction \\
\hline
Loureiro et al. \cite{loureiro2018exploring} & R², RMSE, MAPE, MAE, MSE & Private company data & Deep learning for fashion industry sales forecasting \\
\hline
Lo et al. \cite{lo2019dressing} & MSE, CSE & Lookbook dataset & Social media-based outfit popularity prediction \\
\hline
GeoStyle \cite{mall2019geostyle} & MAE, MAPE & StreetStyle and Flickr 100M & Spatial and temporal trend analysis of fashion attributes \\
\hline
KERN \cite{ma2020knowledge} & MAE, MAPE & Fashion Trend Dataset (FIT) & Integrating domain knowledge with recurrent neural networks \\
\hline
Chakraborty et al. \cite{chakraborty2020predicting} & MSE & Instagram posts of New York Fashion Week (NYFW)-19 & Fashion week social media analysis for trend prediction \\
\hline
Neo-Fashion \cite{zhao2021neo} & mAP & Neo-Fashion Dataset & Comprehensive fashion season forecasting beyond individual elements \\
\hline
Giri et al. \cite{giri2022deep} & RMSE, MAE, CA, F1, AUC & European fashion retailer data & Image-based sales profile prediction for new fashion products \\
\hline
GTM-Transformer \cite{skenderi2024well} & WAPE, MAE, Time Series, ERP & VISUELLE & Exogenous knowledge integration for new product sales forecasting \\
\hline
Sousa et al. \cite{censored} & MAE, MAPE, RMSE, R² & European fashion retailer & Two-stage censored demand forecasting for new products \\
\hline
\end{tabular}
\end{table*}

\section{Future Work and Considerations}\label{sec7}
This study focuses on some less emphasized but potentially important areas within the fashion industry. But a more detailed analysis is necessary to look at the psychological factors and changing dynamics of the needs and wants of Generation Z. Fashion is transient, so newness stands at the risk of obsolescence. Besides, our study did not cover the heterogeneity in people's tastes, which varies because of differences in gender stereotypes and cultural influences. Fashion technology needs to adjust and make room for emerging gender identities. The cultural impact on fashion will have an effect on everything, starting from aesthetic preferences to cultural artifacts, rituals, and appearance-based social norms. Clothes almost all the time are a symbol of heritage, a reflection of religious beliefs, or a manifestation of ethnic identity. Three of these elements will end up creating the identity of the person wearing it, within the context of one cultural group, be it local or national. Thus, there needs to be a focus on that as well. 

 The topic of “fashion" has not historically been taken seriously in discussions in academia. But for many people, especially those who wear their identities on their sleeves, the way they present their gender has a profound effect on what they feel comfortable wearing. Our literature reviews and structure in the future will be radically revised to be more inclusive of all individuals in our target audience. Additionally, another vital step going forward is promoting cultural sensitivity and diversity within fashion aesthetics. This includes recognizing and celebrating cultural influence in design as well as avoiding cultural appropriation, in order to create a more inclusive and respectful fashion industry; one that weaves a narrative of respect and appreciation for one another. Future applications could mean the creation of gender-free fashion design and style; a means for an individual to reach with expression.

\section{Conclusion}\label{sec8}
Our survey synthesized developments across aesthetics, personalization, virtual try-on, and fashion forecasting, connecting these domains to the practical needs of recommender systems. We highlighted how aesthetic modeling and cross-domain visual understanding can inform compatibility and ranking, how personalization leverages preference learning to tailor retrieval and generation, how virtual try-on integrates with detection/segmentation and retrieval to enhance user decision-making, and how forecasting informs trend-aware recommendations and inventory planning. We outlined shared datasets and evaluation metrics and identified open challenges for building cohesive, user-centric, AI-driven fashion platforms.

The cross-domain integration analysis reveals significant opportunities for advancing fashion AI through unified approaches that leverage the strengths of each domain while addressing their individual limitations. Future research should prioritize the development of integrated frameworks that can effectively combine aesthetic assessment, personalized recommendations, realistic virtual try-on, and trend-aware forecasting to create truly intelligent fashion systems that serve user needs across the entire fashion lifecycle.

\section*{Declarations}

\subsection*{Ethical Approval}
Not applicable.

\subsection*{Consent to Participate}
Not applicable.

\subsection*{Consent to Publish}
Not applicable.

\subsection*{Data Availability}
Not applicable.

\subsection*{Authors Contributions}
Both authors contributed to the study conception and design. The first author conducted the survey and analysis. The second author supervised the project. Both authors read and approved the final manuscript.

\subsection*{Funding}
The authors did not receive support from any organization for the submitted work.

\subsection*{Competing Interests}
The authors have no relevant financial or non-financial interests to disclose.

\section{Appendix A: Literature Search Methodology}
This survey employed a systematic literature review approach to identify relevant research works in AI fashion technologies. The search was conducted using Google Scholar as the primary search engine. We used combinations of keywords including “artificial intelligence," “machine learning," “deep learning," “fashion," “clothing," “apparel," “personalization," “recommendation," “virtual try-on," “aesthetic," “style," and “forecasting.“ Publications were limited to the period 2013-2025 to capture both foundational works and recent advancements.

\bibliography{sn-bibliography}

\end{document}